\documentclass[10pt,twocolumn,letterpaper]{article}

\usepackage[pagenumbers]{cvpr} % To force page numbers, e.g. for an arXiv version
\usepackage[accsupp]{axessibility}

\makeatletter
\@namedef{ver@everyshi.sty}{}
\makeatother 

\usepackage{booktabs}

\usepackage{mathptmx}
\usepackage{graphicx}
\usepackage{amsmath}
\usepackage{amsthm}
\usepackage{amssymb} 
\usepackage{pifont}% http://ctan.org/pkg/pifont
\newcommand{\cmark}{\ding{51}}%
\newcommand{\xmark}{\ding{55}}%

\usepackage{wrapfig}

\usepackage[pagebackref=true,breaklinks=true,colorlinks,bookmarks=false]{hyperref}

\usepackage[capitalize]{cleveref}
\crefname{section}{Sec.}{Secs.}
\Crefname{section}{Section}{Sections}
\Crefname{table}{Table}{Tables}
\crefname{table}{Tab.}{Tabs.}

\def\cvprPaperID{229} % *** Enter the CVPR Paper ID here
\def\confName{CVPR}
\def\confYear{2022}

\usepackage{xcolor}
\usepackage{tikz}
\usepackage{changepage}
\usepackage{tikz-3dplot}
\usepackage{tikzscale}
\usetikzlibrary{plotmarks}
\usepackage{pgfplots}
\usetikzlibrary{tikzmark,calc} 
\usetikzlibrary{decorations.pathreplacing, decorations.markings}
  
\usepackage[capbesideposition=right]{floatrow}

\xdefinecolor{navyblue}{RGB}{19 41 75}%{0.14510 0.29020 0.64706} 
\xdefinecolor{carolinablue}{RGB}{123 175 212}
\xdefinecolor{cb}{RGB}{123 175 212}
\xdefinecolor{flatgrey}{RGB}{162 170 173}
\xdefinecolor{matcha}{RGB}{183 224 176}

\xdefinecolor{myyellow}{RGB}{255, 239, 159}
\xdefinecolor{myblue}{RGB}{181, 222, 255}
\xdefinecolor{mypurple}{RGB}{202, 184, 255}

\newcommand{\dashedLayer}[6]{
			\def\a{#1} % Used to distinguish input resolution for current layer.
			\def\b{0.02}
			\def\c{#2} % Width of the cube to distinguish number of input channels for current layer.
			\def\t{#3} % X offset for current layer.
			\def\d{#4} % Y offset for current layer.

			\draw[line width=0.15mm, dash pattern=on 2.25pt off 1.8pt](\c+\t,0,\d) -- (\c+\t,\a,\d) -- (\t,\a,\d);                                                      % back plane
			\draw[line width=0.15mm, dash pattern=on 2.25pt off 1.8pt](\t,0,\a+\d) -- (\c+\t,0,\a+\d) node[midway,below] {#6} -- (\c+\t,\a,\a+\d) -- (\t,\a,\a+\d) -- (\t,0,\a+\d); % front plane 
			\draw[line width=0.15mm, dash pattern=on 2.25pt off 1.8pt](\c+\t,0,\d) -- (\c+\t,0,\a+\d);
			\draw[line width=0.15mm, dash pattern=on 2.25pt off 1.8pt](\c+\t,\a,\d) -- (\c+\t,\a,\a+\d);
			\draw[line width=0.15mm, dash pattern=on 2.25pt off 1.8pt](\t,\a,\d) -- (\t,\a,\a+\d);
			
			\draw[line width=0.15mm] (\c+\t,0,\d) -- (\c+\t,\a,\d);
			\draw[line width=0.15mm] (\c+\t,0,\d) -- (\c+\t,0,\a+\d);
			\draw[line width=0.15mm] (\c+\t,\a,\d) -- (\c+\t,\a,\a+\d);
			\draw[line width=0.15mm] (\c+\t,0,\a+\d) -- (\c+\t,\a,\a+\d);
			
			\filldraw[#5] (\t+\b,\b,\a+\d) -- (\c+\t-\b,\b,\a+\d) -- (\c+\t-\b,\a-\b,\a+\d) -- (\t+\b,\a-\b,\a+\d) -- (\t+\b,\b,\a+\d); % front plane
			\filldraw[#5] (\t+\b,\a,\a-\b+\d) -- (\c+\t-\b,\a,\a-\b+\d) -- (\c+\t-\b,\a,\b+\d) -- (\t+\b,\a,\b+\d);
			\ifthenelse {\equal{#5} {}}
			{} % Do not draw colored slice if #4 is blank.
			{\filldraw[#5] (\c+\t,\b,\a-\b+\d) -- (\c+\t,\b,\b+\d) -- (\c+\t,\a-\b,\b+\d) -- (\c+\t,\a-\b,\a-\b+\d);} % Else, draw a colored slice.
		}

\newcommand{\networkLayer}[6]{
			\def\a{#1} % Used to distinguish input resolution for current layer.
			\def\b{0.02}
			\def\c{#2} % Width of the cube to distinguish number of input channels for current layer.
			\def\t{#3} % X offset for current layer.
			\def\d{#4} % Y offset for current layer.

			\draw[line width=0.15mm](\c+\t,0,\d) -- (\c+\t,\a,\d) -- (\t,\a,\d);                                                      % back plane
			\draw[line width=0.15mm](\t,0,\a+\d) -- (\c+\t,0,\a+\d) node[midway,below] {#6} -- (\c+\t,\a,\a+\d) -- (\t,\a,\a+\d) -- (\t,0,\a+\d); % front plane
			\draw[line width=0.15mm](\c+\t,0,\d) -- (\c+\t,0,\a+\d);
			\draw[line width=0.15mm](\c+\t,\a,\d) -- (\c+\t,\a,\a+\d);
			\draw[line width=0.15mm](\t,\a,\d) -- (\t,\a,\a+\d);

			\filldraw[#5] (\t+\b,\b,\a+\d) -- (\c+\t-\b,\b,\a+\d) -- (\c+\t-\b,\a-\b,\a+\d) -- (\t+\b,\a-\b,\a+\d) -- (\t+\b,\b,\a+\d); % front plane
			\filldraw[#5] (\t+\b,\a,\a-\b+\d) -- (\c+\t-\b,\a,\a-\b+\d) -- (\c+\t-\b,\a,\b+\d) -- (\t+\b,\a,\b+\d);
			\ifthenelse {\equal{#5} {}}
			{} % Do not draw colored slice if #4 is blank.
			{\filldraw[#5] (\c+\t,\b,\a-\b+\d) -- (\c+\t,\b,\b+\d) -- (\c+\t,\a-\b,\b+\d) -- (\c+\t,\a-\b,\a-\b+\d);} % Else, draw a colored slice.
		}
		
\usepackage[linesnumbered,ruled,vlined]{algorithm2e}

\SetCommentSty{mycommfont}
\SetKwInput{Input}{Input}                
\SetKwInput{Output}{Output}         
\SetKwInput{Dataset}{Dataset}     
\SetKwInput{Settings}{Settings}      
\SetKwInput{Initialization}{Initialization}      
\SetAlgorithmName{Alg.}{Alg.}{List of Algorithms} 
\makeatletter
\newcommand{\algorithmfootnote}[2][\footnotesize]{%
  \let\old@algocf@finish\@algocf@finish% Store algorithm finish macro
  \def\@algocf@finish{\old@algocf@finish% Update finish macro to insert "footnote"
    \leavevmode\rlap{\begin{minipage}{\linewidth}
    #1#2
    \end{minipage}}%
  }%
}
\makeatother

\usepackage{floatrow}
\usepackage{caption} 
\usepackage{subcaption}
\usepackage{makecell}
\usepackage{booktabs}
\usepackage{multirow}

\newcolumntype{P}[1]{>{\centering\arraybackslash}p{#1}}

\usepackage{hyperref} 
  
\newtheoremstyle{bfnote}%
  {}{}
  {\itshape}{}
  {\bfseries}{.}
  { }{\thmname{#1}\thmnumber{ #2}\thmnote{ (#3)}}
\theoremstyle{bfnote}
\newtheorem{theorem}{Theorem}

\newtheorem{definition}{Definition}%[section]
\newtheorem*{notation*}{Notation}

\newtheorem{app_theorem}{Theorem}[section]

\usepackage{url}

\newcommand\blfootnote[1]{%
  \begingroup
  \renewcommand\thefootnote{}\footnote{#1}%
  \addtocounter{footnote}{-1}%
  \endgroup
}

\definecolor{olive}{rgb}{0.42, 0.56, 0.14}

\begin{document}

%%%%%%%%% TITLE %%%%%%%%%
%\title{\texttt{D$^2$-SONATA}: Deep Decomposition for Stochastic Normal-Abnormal Transport}
\title{Deep Decomposition for Stochastic Normal-Abnormal Transport}

\author{Peirong Liu\textsuperscript{1} \quad Yueh Lee\textsuperscript{2} \quad Stephen Aylward\textsuperscript{3} \quad Marc Niethammer\textsuperscript{1} \vspace{0.3cm} \\  
\textsuperscript{1}Department of Computer Science, University of North Carolina at Chapel Hill, Chapel Hill, USA\\ \textsuperscript{2}Department of Radiology, University of North Carolina at Chapel Hill, Chapel Hill, USA\\ 
\textsuperscript{3}Kitware Inc., New York, USA \\ \vspace{0.1cm}
{\tt\small \{peirong, mn\}@cs.unc.edu} \quad {\tt\small stephen.aylward@kitware.com} \quad {\tt\small yueh\_lee@med.unc.edu}
} 

\maketitle

%%%%%%%%%%%%%%%%%%%%%%%%%

\begin{abstract}
\vspace{-0.2cm}
  Advection-diffusion equations describe a large family of natural transport processes, e.g., fluid flow, heat transfer, and wind transport. They are also used for optical flow and perfusion imaging computations. We develop a machine learning model, \texttt{D$^2$-SONATA}, built upon a stochastic advection-diffusion equation, which predicts the velocity and diffusion fields that drive 2D/3D image time-series of transport. In particular, our proposed model incorporates a model of transport atypicality, which isolates abnormal differences between expected normal transport behavior and the observed transport. In a medical context such a normal-abnormal decomposition can be used, for example, to quantify pathologies. Specifically, our model identifies the advection and diffusion contributions from the transport time-series and simultaneously predicts an anomaly value field to provide a decomposition into normal and abnormal advection and diffusion behavior. To achieve improved estimation performance for the velocity and diffusion-tensor fields underlying the advection-diffusion process and for the estimation of the anomaly fields, we create a 2D/3D anomaly-encoded advection-diffusion simulator, which allows for supervised learning. We further apply our model on a brain perfusion dataset from ischemic stroke patients via transfer learning. Extensive comparisons demonstrate that our model successfully distinguishes stroke lesions (abnormal) from normal brain regions, while reconstructing the underlying velocity and diffusion tensor fields.%\footnote{Our code and our simulated dataset will be publicly available.}
\blfootnote{This work was supported by the National Institutes of Health (NIH) under award number NIH 2R42NS086295 and NIH R21NS125369.}
\end{abstract}

%\vspace{-0.35cm}

\vspace{-0.2cm}

\section{Introduction}
\label{sec: intro}  
\vspace{-0.15cm}
Partial differential equations (PDEs) are used to describe many transport phenomena, e.g., fluid dynamics, heat conduction, and wind dynamics~\cite{emmanuel2018}. However, it is expensive to numerically solve PDEs especially in high spatial dimensions and across large timescales~\cite{bar2019pde}. Recent deep learning approaches have made data-driven solutions to PDEs possible~\cite{weinan2018ritz,pinn2019,smith2020eikonet,lu2020deeponet,bhattacharya2020reduction,nelsen2020random,sitzmann2020implicit,li2020fourier}. 

While the aforementioned deep learning approaches help to speed-up PDE solutions, we are instead interested in \emph{inverse} PDE problems~\cite{lieberman2010invpde,tartakovsky2020pinn}. Specifically, our goal is to estimate the spatially varying velocity \emph{and} diffusion \emph{tensor} fields (referred to as advection-diffusion parameters throughout this manuscript) for general advection-diffusion PDEs. Limited work on estimating the parameter fields of advection-diffusion equations exists. Tartakovsky et al.~\cite{tartakovsky2020pinn} use a deep neural network (DNN) to estimate 2D diffusion fields from diffusion PDEs. B\'ezenac et al.~\cite{emmanuel2018} learn velocity and diffusion fields of advection-diffusion PDEs by DNNs in 2D. Koundal et al.~\cite{koundal2020omt} use optimal mass transport combined with a spatially constant diffusion. Optimization approaches were proposed~\cite{liu2020piano,liu2021tmi,zhang2021qtm} to estimate the advection-diffusion parameters of an advection-diffusion equation in 3D. Though promising, the numerical optimization approach is time-consuming, especially when dealing with large datasets. Moreover, the methods above assume isotropic diffusion (i.e., they do not estimate more general diffusion tensors), which may be insufficient to accurately model complex materials (e.g., anisotropic porous media, brain tissue) where diffusion is mostly anisotropic.

The \texttt{YETI} approach by Liu et al.~\cite{liu2021yeti} is a deep learning framework to estimate advection-diffusion parameters from transport image time-series. It addresses some key challenges regarding identifiability (i.e., whether the resulting transport should be attributed to advection or diffusion) and physical constraints (e.g., vector fields should be divergence-free for fluid flow, diffusion tensors should be positive semi-definite (PSD)). Applied on brain perfusion images for stroke lesion detection, \texttt{YETI} achieved clear improvements over existing approaches. However, there is still room for improvement. First, while \texttt{YETI} estimates the advection-diffusion parameters well, it would be useful to quantify the inherent uncertainty in the resulting solutions instead of interpreting the results deterministically. This could be especially important  when estimating advection-diffusion parameters from abnormal transport processes, e.g., brain perfusion processes for a stroke patient. Second, \texttt{YETI} predicts advection-diffusion parameters without knowledge of regional anomalies. Hence, additional post-processing is required for lesion detection in real perfusion analysis applications. Furthermore, \texttt{YETI} uses pre-training based on a simulation dataset which only considers normal transport; fine-tuning is then performed using real perfusion data including stroke lesions. This simulation and fine-tuning approach may result in pre-training biased toward normal parameterizations, as abnormal transport data is not simulated. (See the comparisons in Sec.~\ref{exp: 2d_demo}-\ref{exp: ixi}.) 

We introduce, \texttt{D$^2$-SONATA} ({\bf D}eep {\bf D}ecomposition for {\bf S}t{\bf o}chastic {\bf N}ormal-{\bf A}bnormal {\bf T}r{\bf a}nsport), a new deep-learning-based stochastic model designed to predict the underlying advection-diffusion parameters from observed transport time-series with potential anomalies, in both 2D and 3D.\footnote{Perfusion imaging is the motivating application behind our approach. However, our approach is generally applicable to parameter estimation for any process governed by an advection-diffusion equation.} Our main contributions are threefold: 
\\ \vspace{-0.62cm}

\begin{itemize}
\vspace{-0.1cm}
 \item[1)] {\it A learning-based stochastic advection-diffusion model}. \texttt{D$^2$-SONATA} models advection-diffusion processes as a stochastic system. Given a transport process, this stochastic model permits not only reconstructing the transport dynamics with its underlying advection-diffusion parameters, but also captures the epistemic uncertainties with Brownian motion.\\ \vspace{-0.7cm}    
\item[2)] {\it Representation theorems for anomaly-decomposed divergence-free vector fields and symmetric PSD tensor fields}. Our estimates are grounded in theorems ensuring realistic constraints on the learned advection-diffusion parameter fields \emph{by construction}. The estimation automatically provides a decomposition based on an anomaly value field and the ``anomaly-free'' advection-diffusion parameters. This provides insights into both the anomaly patterns and what should locally be considered normal advection-diffusion parameters.  \\ \vspace{-0.7cm}
\item[3)] {\it 2D/3D normal-abnormal advection-diffusion dataset.} We develop a simulator for quasi-realistic advection-diffusion that can be used for supervised model pre-training based on the advection-diffusion parameters. Importantly, the simulator is able to generate velocity vector and diffusion tensor fields and applies artificial anomalies. We show that such simulated data boosts model performance for advection-diffusion parameter prediction, in particular for abnormal transport.  
\end{itemize}

\vspace{-0.2cm}

\section{Related work}
\label{sec: related_work}
\vspace{-0.15cm}
\paragraph{Neural Differential Equations and Optical Flow} 
Significant developments in deep learning have recently led to an explosive growth of deep-learning-based solutions for PDEs. This stream of work either directly models the solution via DNNs~\cite{weinan2018ritz,pinn2019,smith2020eikonet} or learns mesh-free, infinite-dimensional operators using DNNs \cite{lu2020deeponet,bhattacharya2020reduction,nelsen2020random,sitzmann2020implicit,li2020fourier}. 

While the above methods focus on solving PDEs, i.e., the \emph{forward} problem, we are interested in solving \emph{inverse} PDE problems~\cite{lieberman2010invpde,tartakovsky2020pinn}. In the context of advection equations such inverse problems have been extensively studied for  optical flow and general image registration~\cite{horn1980optflow,sun2010optflow}, where the parameter to be estimated is a deformation or a velocity vector field~\cite{horn1981determining,borzi2003optimal,modersitzki2004_numerical,beg2005computing,hart2009optimal}. DNN solutions for the fast prediction of these vector fields have also been studied~\cite{flownet,yang2017quicksilver,balakrishnan2019voxelmorph,shen2019networks}. In contrast, our goal is the estimation of the parameters for more general advection-diffusion PDEs: specifically, their associated velocity vector and diffusion tensor fields. Further, while optical flow and registration approaches most commonly deal with image pairs, our parameter estimation for advection-diffusion PDEs is based on image \emph{time-series}  acquired over multiple time points.

\vspace{-0.25 cm}
\paragraph{Perfusion Imaging}
\label{sec: perfusion} 
Perfusion imaging measures the blood flow through parenchyma by serial imaging~\cite{fieselmann2011sig2ctc}. Common perfusion measurement techniques include injecting an intravascular tracer, e.g., Dynamic Susceptibility Contrast-enhanced (DSC) and Dynamic Contrast-Enhanced (DCE) magnetic resonance (MR) perfusion~\cite{fieselmann2011sig2ctc,demeestere2020stroke,liu2021tmi}, using magnetically-labeled arterial blood water protons as an endogenous tracer (Arterial spin labeling (ASL))~\cite{asl2010}, or using positron emission tomography (PET)~\cite{pet2011}. Derived quantitative measures help clinical diagnosis and decision-making. %, e.g., to assess acute strokes and brain tumors. 
So far, the mainstream approach for quantitative perfusion measures is to use tracer kinetic models to estimate hemodynamic parameters and to obtain 3D perfusion parameter maps~\cite{fieselmann2011sig2ctc,mouridsen2006aif}. However, substantial differences exist in perfusion parameter maps across institutions, mainly caused by different arterial input function (AIF) selection procedures, deconvolution techniques, and interpretations of perfusion parameters \cite{mouridsen2006aif,schmainda2019a,schmainda2019b}. Moreover, these approaches are performed on individual voxels, disregarding spatial dependencies of the injected tracer's dynamics. 

Works to fit tracer transport via PDEs exist, where the observed tracer concentration time-series is assumed to reflect the blood flow within the vessels (advection) while diffusion captures the movements of freely-diffusive tracer within capillaries and the macroscopic effect of capillary transport~\cite{cookson2014spatial,harabis2013dilution,strouthos2010dilution,cookson2014spatial,liu2021tmi}. However, these works assume that both velocity and diffusion are {\it constant} over the entire domain, which is unrealistic in real tissue. Zhou et al.~\cite{zhou2018vector,zhou2021qtm} and Liu et al.~\cite{liu2020piano,liu2021tmi} proposed to model the perfusion process via a transport model using numerical optimization. However, Zhou et al.~\cite{zhou2018vector,zhou2021qtm} assume the diffusion process is negligible; therefore, only the velocity field is estimated similar to optical flow~\cite{horn1980optflow,borzi2003optimal,niethammer2009optimal}. Liu et al.~\cite{liu2020piano,liu2021tmi} and Zhang et al.~\cite{zhang2021qtm} estimate both the spatial-varying velocity and diffusion fields, yet modeling the diffusion as a scalar field, which cannot express the diffusion anisotropy.

Liu et al.~\cite{liu2021yeti} propose to predict the perfusion process via an advection-diffusion model with {\it{intrinsically constructed}} divergence-free velocity vector fields and symmetric positive-semi-definite (PSD) diffusion {\it{tensor}} fields, using a deep-learning approach. This work allows for modeling diffusion anisotropy via tensors, and significantly reduces the inference time by resorting to deep learning approaches. However, regional anomalies are not explicitly modeled. Hence, additional post-processing steps, such as lesion detection, are required in perfusion analysis applications. In contrast, \texttt{D$^2$-SONATA} not only predicts the spatial dependency of advection-diffusion parameters in a stochastic manner, but it also decomposes transport into its normal and abnormal parts and can thereby provide additional insights into existing patterns of anomaly.

\section{Method}
\label{sec: method}

%We propose three contributions over \texttt{YETI}~\cite{liu2021yeti}, namely ... To make the manuscript self-contained, we first give the necessary theoretical background on the original \texttt{YETI}~\cite{liu2021yeti} (Sec.~\ref{sec: yeti}).

%%%%%%%%%%%%%%%%%%%%

\vspace{-0.1cm}
\subsection{Problem Setup}
\label{sec: setup}

\vspace{-0.1cm}
Let $C = C({\mathbf{x}}, t)$ denote the mass concentration at location ${\mathbf{x}}$ in a bounded domain $\Omega\subset \mathbb{R}^d\,(d = 2,\,3)$, at time $t \in [0,\, T]$. The local mass concentration changes of an advection-diffusion process can be modeled as:
\vspace{-0.15cm}
\begin{align}
\label{eq: adv_diff} 
\frac{\partial C}{\partial t} & = - {\mathbf{V}}\cdot\nabla C + \nabla \cdot \left({\mathbf{D}}\, \nabla C\right) + \sigma \partial W_t  \\ \vspace{-0.15cm}
& = \underbrace{- {(A \diamond \overline{\mathbf{V}}})\cdot\nabla C}_{\substack{\text{Anomaly-decomposed}\\\text{Incompressible flow}}} + \underbrace{\nabla \cdot \left( (A \circ {\overline{\mathbf{D}}}) \, \nabla C\right)}_{\substack{\text{Anomaly-decomposed}\\\text{PSD Diffusion}}} + \underbrace{\sigma \partial W_t}_{\substack{\text{Model}\\\text{Uncertainty}}}, \nonumber   
\vspace{-0.5cm}
\end{align}
with specified boundary conditions (B.C.). The advection term captures the transport related to the flow of the fluid and the diffusion is driven by the gradient of mass concentration. The spatially-varying velocity field ${\mathbf{V}}$ (${\mathbf{V}}(\mathbf{x}) \in \mathbb{R}^d$) and diffusion tensor field ${\mathbf{D}}$ (${\mathbf{D}}(\mathbf{x}) \in \mathbb{R}^{d\times d}$) describe the advection and diffusion. Eq.~(\ref{eq: adv_diff}) contains the common assumption that fluids show negligible density variation in practice~\cite{kim19deepfluid,liu2021yeti}, which means ${\mathbf{V}}$ has zero divergence (i.e., is divergence-free).\footnote{Note when $\mathbf{D} \to 0$, Eq.~(\ref{eq: adv_diff}) is an advection equation, which is the basis for many variational optical flow or image registration methods~\cite{horn1981determining,borzi2003optimal,modersitzki2004_numerical,beg2005computing,hart2009optimal,deepflow2013,flownet,yang2017quicksilver,balakrishnan2019voxelmorph,shen2019networks}. If density variations should be modeled one can simply replace $\mathbf{V}\cdot\nabla C$ with $\nabla \cdot (C \, \mathbf{V})$. (See more details in Supp.~\ref{app: numericals}.)} As in \cite{liu2021yeti}, we model diffusion tensors ${\mathbf{D}(x)}$ as symmetric PSD, to capture the predominant diffusion direction and diffusion anisotropy~\cite{niethammer2006dti}. 

In contrast to existing approaches we explicitly model abnormalities via an anomaly value field $A$ ($A(x) \in \mathbb{R}_{(0,\, 1]}$, higher $A(x)$ means closer to normal), which modulates both the velocity and diffusion fields. Specifically, we assume that the velocity field can be decomposed as $\mathbf{V}=A\diamond\overline{\mathbf{V}}$ and the diffusion tensor field as $\mathbf{D}=A\circ\overline{\mathbf{D}}$, where $\overline{\mathbf{V}}$ and $\overline{\mathbf{D}}$ denote the ``anomaly-free'' velocity and diffusion tensor fields respectively. Here, $\diamond$ and $\circ$ denote the chosen interactions between the ``anomaly-free'' velocity and diffusion tensor fields and the anomaly value field respectively (see the definition in Eq.~(\ref{eq: relation})). This interaction can be chosen in different ways. See Sec.~\ref{sec: repre} (Theorems~\ref{thm: repre_v} \&~\ref{thm: repre_psd} and Definition~\ref{def: relation}) for details on the anomaly-decomposed representations (and how $\diamond$ and $\circ$ are operationalized) which assure that the decompositions can express divergence-free velocity fields and PSD diffusion tensor fields.

%\begin{definition}[Brownian motion] A stochastic process ($W_t$) such that (1) $W_0 = 0$; (2) $(W_t  - W_s) \sim \mathcal{N}(0, t-s), ~\forall t \geq s \geq 0$; (3) For all disjoint time interval pairs $[t_1, t_2]$, $[t_3, t_4]$ $(t_1< t_2 \leq t_3 \leq t_4)$, the increments $W_{t_4} - W_{t_3}$ and $W_{t_2} - W_{t_1}$ are independent random variables.
%\end{definition}

In addition, we model the advection-diffusion process via a stochastic PDE (SPDE), where $\sigma$ $(\sigma(\mathbf{x}) \in \mathbb{R})$ denotes the variance of the Brownian motion 
$W_t\, (W(\mathbf{x},\, t)\in \mathbb{R})$~\cite{Guerra2009brownian} and represents the epistemic uncertainty for the dynamical system. With this additional stochastic term, the existence and uniqueness of the solution to Eq.~(\ref{eq: adv_diff}) still holds~\cite{oksendal2010stochastic,Guerra2009brownian,Demian2014probreg}. % (See complete proof for Theorem~\ref{thm: spde} in Supp.~\ref{app: proof_spde}.). 

%\begin{theorem}[Existence and uniqueness of SPDE]If the coefficients of the stochastic partial differential equation Eq.~(\ref{eq: adv_diff}) with initial condition, satisfy the spatially-varying Lipschitz condition
%\vspace{-0.15cm}
%\begin{align}
%& |\mathbf{V}(\mathbf{x}_1) - \mathbf{V}(\mathbf{x}_2)|^2 + |\mathbf{D}(\mathbf{x}_1) - \mathbf{D}(\mathbf{x}_2)|^2 \nonumber \\
%& \qquad\qquad + |\sigma(\mathbf{x}_1) - \sigma (\mathbf{x}_2)|^2 \leq K|\mathbf{x}_1 - \mathbf{x}_2|^2,
%\vspace{-0.15cm}
%\end{align}
%and the spatial growth condition
%\vspace{-0.1cm}
%\begin{equation}
%|\mathbf{V}(\mathbf{x})|^2 +|\mathbf{D}(\mathbf{x})|^2 + |\sigma(\mathbf{x})|^2 \leq K ( 1 + \mathbf{x}^2 ),
%\vspace{-0.1cm}
%\end{equation}
%then there is a continuous adapted solution $C({\mathbf{x}}, t)$ satisfying the $L^2$ bound. Moreover, if $C({\mathbf{x}}, t)$ and $\widetilde{C}({\mathbf{x}}, t)$ are both continuous solutions satisfying the $L^2$ bound, then
%\vspace{-0.1cm}
%\begin{equation}
%P\left(C({\mathbf{x}}, t) = \widetilde{C}({\mathbf{x}}, t) \text{ for all } t \in [0, \, T]\right) = 1.
%\vspace{-0.1cm}
%\end{equation}
%\label{thm: spde}
%\end{theorem}

%%%%%%%%%%%%%%%%%%%%

%\vspace{-0.1cm}
\subsection{Anomaly-decomposed Constraint-free Representation}
\label{sec: repre}
\vspace{-0.15cm}

As discussed in Sec.~\ref{sec: setup}, incompressibility and symmetric PSD-ness are commonly used assumptions for fluid flow and diffusion. In \texttt{YETI}~\cite{liu2021yeti}, divergence-free vectors and PSD tensors are obtained by construction via a suitable parameterization. Hence, when integrated into a deep-learning model no extra losses need to be imposed during training to obtain these properties. However, \texttt{YETI} does not provide an explicit abnormality model for the predicted $\mathbf{V}$ and ${\mathbf{D}}$ fields. \texttt{D$^2$-SONATA}, therefore introduces two representation theorems which allow constructing divergence-free velocity fields and symmetric PSD diffusion tensor fields in such a way that a decomposition into normal and abnormal value fields is obtained.

%%%%%%%%%%%%%%%%%%%%%%%%%%%%%%%%%%%

\vspace{-0.35 cm}
\paragraph{Anomaly-decomposed Divergence-free Velocity Vectors}
\label{sec: repre_v}

B\'ezenac et al.~\cite{emmanuel2018} penalize deviations of predicted velocity fields from zero divergence during training, with the assumption that test time predictions will be \emph{approximately} divergence free. Kim et al.~\cite{kim19deepfluid} parametrize velocity vectors via the curl of vector fields, yet do not account for the boundary conditions that need to be imposed in bounded domain scenarios~\cite{dubois1990div_free,amrouche1998div_free,maria2003hhd,amrouche2013div_free}. Liu et al.~\cite{liu2021yeti} propose to predict a parameterization of velocity fields $\mathbf{V}$ which is by construction divergence-free; however, abnormal regions are not explicitly modeled for $\mathbf{V}$ and extra post-processing steps are required for applications targeted at anomaly detection. 
Therefore, our goal is a representation strategy for $\mathbf{V}$ on a domain $\Omega\subset \mathbb{R}^d\,(d = 2,\,3)$ with a smooth boundary such that 
(1) $\mathbf{V}$ is divergence-free {\it{by construction}}; 
(2) {\it{Any}} divergence-free $\mathbf{V}$ can be represented; 
(3) $\mathbf{V}$ can be decomposed into an anomaly value field $A$ which characterizes the abnormal fields, and $\overline{\mathbf{V}}$ which refers to the corresponding normal value field without anomaly. 

\vspace{-0.1cm} 
\begin{theorem}[Anomaly-decomposed Divergence-free Vector Representation]For any vector field $\mathbf{V} \in L^p(\Omega)^d$ and scalar field $A$ in $\mathbb{R}_{(0,\, 1]}(\Omega)$ on a bounded domain $\Omega\subset\mathbb{R}^d$ with smooth boundary $\partial \Omega$. If $\mathbf{V}$ satisfies $\nabla \cdot \mathbf{V} = 0$, there exists a potential $\boldsymbol{\Psi}$ in $L^p(\Omega)^{\alpha}$ ($\alpha = 1(3)$ when $d = 2(3)$):
\vspace{-0.1cm} 
\begin{equation}
\mathbf{V} = \nabla \times (A \, \boldsymbol{\Psi}), \quad (A \, \boldsymbol{\Psi}) \cdot \mathbf{n}\big|_{\partial \Omega} = 0.
\label{eq: repre_v} 
\vspace{-0.1cm}
\end{equation}
Conversely, for any $A \in \mathbb{R}_{(0,\, 1]}(\Omega)$, $\boldsymbol{\Psi}\in L^p(\Omega)^{\alpha}$, $\nabla \cdot \mathbf{V} =\nabla \cdot (\nabla \times (A \, \boldsymbol{\Psi}))= 0$. 
\label{thm: repre_v}
(See complete proof in Supp.~\ref{app: proof_div_free}.)
\end{theorem}

%%%%%%%%%%%%%%%%%%

\vspace{-0.45 cm}
\paragraph{Anomaly-decomposed Symmetric PSD Diffusion Tensors}
\label{sec: repre_d}

We seek a representation for $\mathbf{D}$ (for a $\mathbf{D}(x)$) such that  
(1) $\mathbf{D}$ is a symmetric PSD tensor {\it{by construction}}; 
(2) {\it any} symmetric PSD $\mathbf{D}$ can be represented;
(3) $\mathbf{D}$ can be decomposed into an anomaly value field $A$ indicating the abnormal patterns, and $\overline{\mathbf{D}}$ which is the corresponding normal value field. We resort to the spectral decomposition theorem and the surjective Lie exponential mapping on $SO(n)$ ($exp:\, \mathfrak{so}(n) \mapsto  SO(n)$, $\mathfrak{so}(n)$ is the group of skew-symmetric matrices, $SO(n)$ is the real orthogonal group)~\cite{lezcano2019spectral,lezcano2019trivializations,liu2021yeti}.

\vspace{-0.1cm} 
\begin{theorem}[Anomaly-decomposed Symmetric PSD Tensor Representation]
For any $n\times n$ symmetric PSD tensor $\mathbf{D}$ and $A \in \mathbb{R}_{(0,\, 1]}(\Omega)$, there exist an upper triangular matrix with zero diagonal entries, $\mathbf{B} \in \mathbb{R}^{\frac{n(n-1)}{2}}$, and a non-negative diagonal matrix, $\boldsymbol{\Lambda} \in SD(n)$, satisfying:
\vspace{-0.15cm} 
\begin{equation}
\mathbf{D} = \mathbf{U}\, (A \, \boldsymbol{\Lambda}) \, \mathbf{U}^T,\quad \mathbf{U} = exp(\mathbf{B} - \mathbf{B}^T) \in SO(n).
\label{eq: repre_psd} 
\vspace{-0.25cm}
\end{equation}
Conversely, for $\forall A \in \mathbb{R}_{(0,\, 1]}(\Omega)$, $\forall \mathbf{B} \in \mathbb{R}^{\frac{n(n-1)}{2}}$, and $\forall \boldsymbol{\Lambda} \in SD(n)$, Eq.~(\ref{eq: repre_psd}) results in a symmetric PSD tensor, $\mathbf{D}$. 
\label{thm: repre_psd}
(See complete proof in Supp.~\ref{app: proof_psd}.)
\end{theorem}

%%%%%%%%%%%%%%%%%%
\vspace{-0.35cm} 
\begin{definition}[``Anomaly-free'' Fields] According to Eq.~(\ref{eq: repre_v}-\ref{eq: repre_psd}), when $A$ equals $1$ over the entire domain ($\Omega$), we denote the corresponding parameters as ``anomaly-free''. For the sake of convenience, we write the relation: 
\vspace{-0.25cm}
\begin{equation}
\mathbf{V} = A \diamond \overline{\mathbf{V}}, \,\, \mathbf{D} = A \circ \overline{\mathbf{D}},
\label{eq: relation}
\vspace{-0.3cm}
\end{equation}
where the overline is used to denote ``anomaly-free'' fields. Obviously, when $A \to 1$, $\mathbf{V},\, \mathbf{D}$ are closer to normal. (See explicit expressions of the $\diamond$, $\circ$ operations in Supp.~\ref{app: proof_div_free} \& \ref{app: proof_psd}.)
\vspace{-0.1cm} 
\label{def: relation}
\end{definition}
In this way, we not only obtain the underlying advection-diffusion parameters ($\mathbf{V}$, $\mathbf{D}$) satisfying the divergence-free and PSD constraints from the observed transport processes \emph{by construction}, but we also gain knowledge as to where the anomalies exist and what the normal patterns of the advection-diffusion parameters are. Note that in our proposed approach we will directly predict $A$, $\overline{\mathbf{V}}$, and $\overline{\mathbf{D}}$. I.e., we will {\it{directly}} work in the decomposed domain rather than predicting $\mathbf{V}$ and $\mathbf{D}$ followed by a decomposition.

%%%%%%%%%%%%%%%%%%%%
\subsection{\texttt{D$^2$-SONATA}: Deep Decomposition for StOchastic Normal-Abnormal TrAnsport}
\label{sec: network} 

\vspace{-0.1cm}
%%%%%%%%%%%%%%%%%%%%%%%%%%%%

\input{sub/method/fw}  

%%%%%%%%%%%%%%%%%%%%%%%%%%%%

Sec.~\ref{sec: repre} described the representation theorems of \texttt{D$^2$-SONATA} for anomaly-encoded divergence-free velocity fields and symmetric PSD diffusion fields. This enables expressing the advection-diffusion parameters with realistic constraints along with the corresponding anomaly patterns \emph{by construction}. In particular, using these representations within a deep network will assure these constraints during training \emph{and} testing. This section introduces  \texttt{D$^2$-SONATA}'s two-stage learning framework, to predict the velocity vector and diffusion tensor fields driving a potentially anomalous observed advection-diffusion process:

\noindent (1) \emph{Physics-informed learning}: reconstructs the anomaly value field ($A$) and ``anomaly-free'' advection-diffusion parameters ($\overline{\mathbf{V}},\, \overline{\mathbf{D}}$), from observed transport time-series. The anomaly-encoded parameters ($\mathbf{V},\, \mathbf{D}$) can then be obtained (Eq.~(\ref{eq: relation})). In this stage, the model is trained on a simulated dataset (Sec.~\ref{exp: 2d_demo}-\ref{exp: ixi}) under the supervision of the ground truth parameters and a corresponding anomaly value field; 

\noindent (2) \emph{Transport-informed learning}: reconstructs the transport concentration time-series, where the ground truth advection-diffusion parameters can be unknown. During this stage, the supervision is imposed by the observed mass transport dynamics, via the integration of the stochastic advection-diffusion PDE in Eq.~(\ref{eq: adv_diff}).

\vspace{-0.35cm}
\paragraph{Patch-based Input Time-series}
\label{sec: input}
Given a time-series $C = \{ C^{t_i} \in \mathbb{R}(\Omega) \big| i = 1,\, 2,\, ...,\, N_{T} \}$, we first extract $32^3$ ($32^2$ for 2D domains) patches randomly (uniform over the space) over the same spatial domain ($\Omega_p\subset\Omega$) at $N_{\text{in}}$ sequential time points, starting from a random $t_i$ ($i \in \{1,\, 2,\,...,\, N_{T}-N_{\text{in}}+1\}$). Each training sample ($C_p$) is denoted as $C_p = \{ C_p^{t_j} \in \mathbb{R}(\Omega_p) \big| j = i,\, ...,\, i+N_{\text{in}}-1 \}$ (Fig.~\ref{fw} (top left)).

%%%%%%%%%%%%%%%%%%%%%%%%%%%%
%%%%%%%%%%%%%%%%%%%%%%%%%%%%

\vspace{-0.35cm}
\paragraph{Physics-informed Learning} 
\label{sec: vd_net}
We use a U-Net~\cite{3dunet2016,2dunet2015}) as the backbone for the parameter prediction network ($\mathcal{F}_{P}$) in this stage (Fig.~\ref{fw} (middle)). A \emph{transport} encoder is first used to encode the input $C_p$ into latent features. The number of the input time points determines the input channels. Two decoders then learn the mappings to the representations of the ``anomaly-free'' parameters ($\overline{\mathbf{V}}$, $\overline{\mathbf{D}}$). Specifically, the $\overline{\mathbf{V}}$-decoder predicts the potential $\widehat{\boldsymbol{\Psi}}$ to represent a divergence-free velocity field $\widehat{\overline{\mathbf{V}}}$ via Theorem~\ref{thm: repre_v}, and the $\overline{\mathbf{D}}$-decoder predicts $\widehat{\mathbf{B}}$ along with $\widehat{\boldsymbol{\Lambda}}$ to represent a symmetric PSD diffusion tensor field $\widehat{\overline{\mathbf{D}}}$ via Theorem~\ref{thm: repre_psd}. Another decoder outputs the predicted anomaly value field ($A$). The resulting $\widehat{A}$,  $\widehat{\overline{\mathbf{V}}}$ and $\widehat{\overline{\mathbf{D}}}$ then allow computing  the final prediction of the estimated velocity ($\widehat{\mathbf{V}}$) and diffusion ($\widehat{\mathbf{D}}$) fields (Eq.~(\ref{eq: relation})). 

The reconstruction loss in this stage is three-fold. It is a supervised loss based on the actual and ``anomaly-free'' advection-diffusion parameters and the anomaly value field, which are all known from simulation:
\vspace{-0.25cm}
\begin{equation}
\begin{aligned}
\hspace{-0.1cm}\mathcal{L}_{\mathbf{VD}} & = \frac{1}{|\Omega_p|}\int_{\Omega_p} \bigg\{ \big\| \overline{\mathbf{V}}- \widehat{\overline{\mathbf{V}}} \big\|_{2} + \big\| \mathbf{V}- \widehat{\mathbf{V}} \big\|_{2} \vspace{-1cm} \\
& \qquad + \big\| \overline{\mathbf{D}}- \widehat{\overline{\mathbf{D}}} \big\|_{\text{F}} + \big\| \mathbf{D} - \widehat{\mathbf{D}} \big\|_{\text{F}} + \vert A - \widehat{A} \vert \bigg\} \, d\mathbf{x}\,,
\label{eq: loss_vd} 
\end{aligned} 
\vspace{-0.4cm}
\end{equation}
%where $\mathbf{V},\,\overline{\mathbf{V}},\, \mathbf{D}\,\overline{\mathbf{D}}$ denote the ground truth parameters,
where the hatted symbols denote the predictions, $\|\cdot\|_{2}$, $\|\cdot\|_{\text{F}}$ the vector 2-norm and tensor Frobenius norm respectively. 

Similar to \texttt{YETI} \cite{liu2021yeti}, we additionally impose supervision on the eigenvectors ($\widehat{\mathbf{U}}$) and eigenvalues ($\widehat{\boldsymbol{\Lambda}}$) of $\widehat{\mathbf{D}}$ to improve the network's ability to capture the anisotropic structure of diffusion tensors. Note $\widehat{\mathbf{U}} = [\widehat{\mathbf{u}}_1,\, \widehat{\mathbf{u}}_2,\, \widehat{\mathbf{u}}_3]$ is an intermediate output from $\widehat{\mathbf{B}}$ via Eq.~(\ref{eq: repre_psd}). 
\vspace{-0.3cm}
\begin{equation} 
\mathcal{L}_{\mathbf{U}\boldsymbol{\Lambda}} = \frac{1}{|\Omega_p|}\int_{\Omega_p} \sum_{i=1}^{3}min\big\{\big\|\mathbf{u}_i \pm \widehat{\mathbf{u}}_i \big\|_{2}\big\}  + \big\| \boldsymbol{\Lambda} - \widehat{\boldsymbol{\Lambda}} \big\|_{\text{F}}\,d\mathbf{x}\,,
\label{eq: loss-lu} 
\vspace{-0.25cm}
\end{equation}
where $min$ addresses the eigenvector sign ambiguities by selecting the minimum between $\|\mathbf{u}_i + \widehat{\mathbf{u}_i}\|_2$ and $\|\mathbf{u}_i - \widehat{\mathbf{u}_i}\|_2$.

The loss for the physics-informed learning stage is
\vspace{-0.25cm}
 \begin{equation}
\mathcal{L}_{\text{Phy}} = \mathcal{L}_{\mathbf{VD}} + w_{\mathbf{U}\boldsymbol{\Lambda}} \mathcal{L}_{\mathbf{U}\boldsymbol{\Lambda}},\quad w_{\mathbf{U}\boldsymbol{\Lambda}} > 0. 
\label{eq: phy}
\vspace{-0.2cm}
\end{equation}

%%%%%%%%%%%%%%%%%%%%%%%%%%%%
%%%%%%%%%%%%%%%%%%%%%%%%%%%%

\vspace{-0.45cm}
\paragraph{Transport-informed Learning}
\label{sec: integration}
In this stage (Fig.~\ref{fw}), the model is trained under the supervision of the transport dynamics. It contains two parts: (1) an uncertainty prediction network ($\mathcal{F}_{U}$); (2) a stochastic advection-diffusion PDE integrator. Specifically, $\mathcal{F}_{U}$ is used to represent the model's epistemic uncertainty, namely $\sigma$ in Eq.~(\ref{eq: adv_diff}). Intuitively, the model's epistemic uncertainty should capture the following: (1) For values within the normal range, the variance of the Brownian motion should be small (low uncertainty). In this case the system state is dominated by the advection-diffusion term. (2) For anomaly regions with parameter values outside the normal range, the variance of the
Brownian motion should be large and the system should  exhibit higher uncertainty~\cite{kong2020sde}. 
Based on this desired property, we design our training loss for model uncertainty as
\vspace{-0.25cm} 
\begin{equation} 
\mathcal{L}_{\sigma} = \frac{1}{|\Omega_p|}\int_{\Omega_p} \left( (1 - A)- \widehat{\sigma} \right)^{2}~d{\mathbf{x}},
\label{eq: loss_sigma}
\vspace{-0.25cm}
\end{equation}
with the prior assumption that uncertainties should be small when the ground truth $A$ is close to the normal value, $1$.

Next, we implement a stochastic advection-diffusion PDE solver to integrate the initial state ($C_{p}^{t_i}$) forward in time to $t_{i+N_{\text{out}}-1}$ via Eq.~(\ref{eq: adv_diff}) (Fig.~\ref{fw} (right)), and to train the model by minimizing the differences between the predicted ($\widehat{C}_p$) and the input ($C_p$) time-series. We follow the patch-based Cauchy B.C. and the spatial discretization scheme described in  Liu et al.~\cite{leveque2002,tartakovsky2020pinn,liu2021yeti}. We also use RK45 for time-integration ($\delta t$) for the concentration prediction $\widehat{C}^{t+\delta t}$. For the stochastic term ($\sigma$) in Eq.~(\ref{eq: adv_diff}), we use the Euler-Maruyama scheme \cite{Kloeden1992Stochastic,kong2020sde}, and the discretized version of the stochastic term in Eq.~(\ref{eq: adv_diff}) can therefore be written as $\sigma \, W / \sqrt{\Delta t}$, where $W \sim \mathcal{N}(0,\, 1)$.\footnote{Note when the temporal resolution ($\Delta t$) is coarse, $\delta t$ should be set smaller than $\Delta t$ (Fig.~\ref{fw} (top right)) to satisfy the Courant-Friedrichs-Lewy (CFL) condition \cite{gottlieb2000ssp,leveque2002} and to allow for stable numerical integration.} (See Supp.~\ref{app: numericals} for details.)
  
%%%%%%%%%%%%%%%%%%%%%%%%%%%%

Given a training patch sample $\{C_p^{t_j} \in \mathbb{R}(\Omega_p)| j = i,\, \ldots,\, i+N_{\text{in}}-1\}$, we compute the collocation concentration loss ($\mathcal{L}_{\text{CC}}$)~\cite{mathieu2016gdl,liu2021yeti}, the mean squared error of the predicted time series at output collocation time
points, to encourage the predicted values to be close to the observed ones. We also use a regularizer ($\mathcal{L}_{\text{SS}}$)~\cite{liu2021yeti} on the gradient fields of each component of  $\widehat{\mathbf{V}},\, \widehat{\mathbf{D}}$ to encourage the predicted parameter fields to be spatially smooth.  

%The complete loss for this transport-informed learning stage is
The complete loss for this stage is thus ($w_{\text{SS}},\, w_{\sigma} > 0$)
\vspace{-0.25cm} 
\begin{equation}
\mathcal{L}_{\text{Trn}} = \mathcal{L}_{\text{CC}} + w_{\text{SS}}\, \mathcal{L}_{\text{SS}} + w_{\sigma} \mathcal{L}_{\sigma}\,. 
\label{eq: lat} 
\vspace{-0.2cm} 
\end{equation}

%%%%%%%%%%%%%%%%%%%%

\section{Experiments}
\label{sec: exp}
\vspace{-0.2cm}
Secs.~\ref{exp: 2d_demo}-\ref{exp: ixi} present results on simulated time-series in 2D/3D. We analyze the individual contribution of \texttt{D$^2$-SONATA}'s components, and compare the improvements achieved to the state of the art approaches. In Sec.~\ref{exp: isles}, we further transfer our model pre-trained on simulated data to real time-series of magnetic resonance (MR) perfusion images from ischemic stroke patients. We demonstrate \texttt{D$^2$-SONATA}'s ability to distinguish stroke lesions from normal brain regions via its predicted anomaly value field $\widehat{A}$ and reconstructed advection-diffusion parameters ($\widehat{\mathbf{V}}$, $\widehat{\mathbf{D}}$). 

\begin{table}[t]
\resizebox{\linewidth}{!}{
\centering
    %\begin{tabular}{P{0.6cm}P{1.3cm}|P{0.6cm}P{0.6cm}P{0.6cm}|P{0.6cm}P{0.6cm}|P{0.6cm}P{0.6cm}P{0.6cm}} 
    \begin{tabular}{ccccccc} 
       \toprule \\[-3ex] 
      % Models & \thead{\normalfont{Advection-diffusion}\\\normalfont{Parameters ($\mathbf{V,\, D}$)}} & \thead{\normalfont{``Anomaly-free''}\\\normalfont{Parameters ($\overline{\mathbf{V}},\, \overline{\mathbf{D}}$)}} & \thead{\normalfont{Anomaly Value}\\\normalfont{Field ($M$)}} & \thead{\normalfont{Uncertainty}\\\normalfont{($\sigma$)}} & Notes \\ [-0.2ex]
      {\bf{Models}} & {\bf{Training Data}} & $\mathbf{V,\, D}$ & $\overline{\mathbf{V}},\, \overline{\mathbf{D}}$ & $A$ & $\sigma$ & {\bf{Type}} \\ [-0.2ex]
     \midrule\\[-3ex]
     %Kinetic~\cite{fieselmann2011sig2ctc} & N/A & \xmark & \xmark & \xmark & \xmark & Per-voxel computation  \\
     %\hline\\[-2.4ex]
     \texttt{PIANO}~\cite{liu2020piano} & N/A & \cmark & \xmark & \xmark & \xmark & Optimization-based  \\
     \hline\\[-2.4ex]
     \texttt{YETI}~\cite{liu2021yeti} & Normal & \cmark & \xmark & \xmark & \xmark & Learning-based  \\
     \hline\\[-2.4ex]
     Anom.-\texttt{YETI} & (Ab-)Normal & \cmark & \xmark & \xmark & \xmark & Learning-based  \\
     \hline\\[-2.4ex]
     w/o-$\sigma$ & (Ab-)Normal & \cmark & \cmark & \cmark & \xmark & Learning-based  \\
     \hline\\[-2.4ex]
     w/o-$A$ & (Ab-)Normal & \cmark & \cmark & \xmark & \cmark & Learning-based  \\
     \hline\\[-2.4ex]
     \texttt{D$^2$-SONATA} & (Ab-)Normal & \cmark & \cmark & \cmark & \cmark & Learning-based  \\

\bottomrule  \\ [-3.2ex] 
	\vspace{-0.7cm}
    \end{tabular}  
    \caption{Model properties comparison.} 
    \label{tab: models}
}

\end{table}

%%%%%%%%%%%%%%%%%%%%%%%%%%%%%%% 
\vspace{-0.15cm}
\subsection{2D (Ab-)Normal Anisotropic Moving Gaussian}
\label{exp: 2d_demo}

\input{sub/exp/fig/2d_demo_plt} 

\begin{figure}[t]
\hspace*{-0.8cm}
	\noindent\resizebox{1.15\textwidth}{!}{
	\begin{tikzpicture} 
	
	%\node at (-6, -1) {Err(C)};
	
	\node at (-6, -2) {\includegraphics[width=0.36\textwidth]{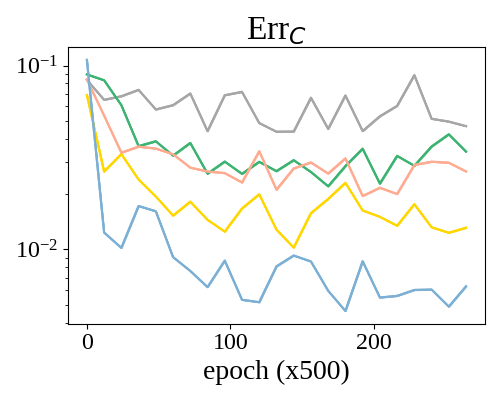}};
	\node at (-3, -2) {\includegraphics[width=0.36\textwidth]{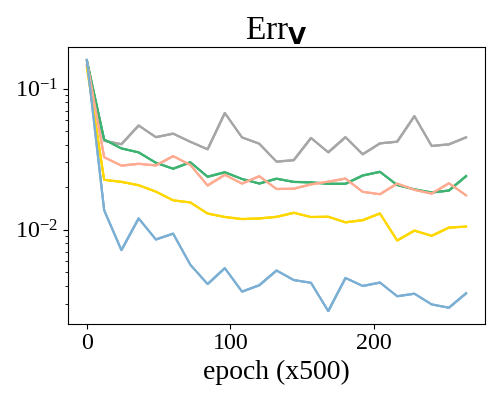}};
	\node at (0, -2) {\includegraphics[width=0.36\textwidth]{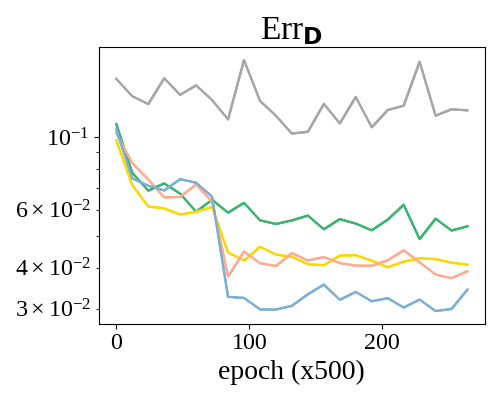}};
	%\node at (3, -2) {\includegraphics[width=0.175\textwidth]{fig/2d_demo/loss/Err_U.png}};
	%\node at (6, -2) {\includegraphics[width=0.175\textwidth]{fig/2d_demo/loss/Err_L.png}};  
	\node at (-3, -3.4) {\includegraphics[width=0.8\textwidth]{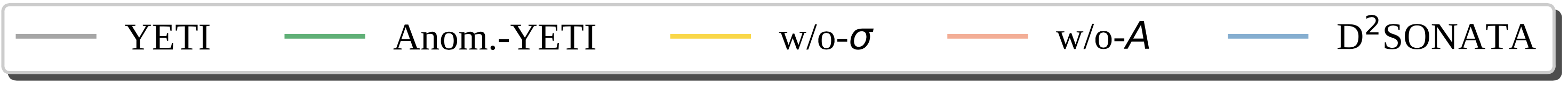}};

	\end{tikzpicture}
	}
	\vspace{-0.8cm}
	\caption{Mean relative absolute error (RAE) of \texttt{YETI}, Anom.-\texttt{YETI}, w/o-$\sigma$, w/o-$A$ and \texttt{D$^2$-SONATA} for 2D ab-(normal) anisotropic moving Gaussian (Sec.~\ref{exp: 2d_demo}). X-axis: training iterations; Y-axis: the average RAE (log scale) of 100 test samples.}
	\label{fig: 2d_demo_loss_comp}
\end{figure}

\vspace{-0.15cm}
\paragraph{Dataset}
\label{sec: 2d_dataset} 

\vspace{-0.1cm}
We simulate 2D advection-diffusion with random anomalies on-the-fly (Fig.~\ref{fig: 2d_demo}). Each sample is a 2D image time-series ($N_T = 40$, $\Delta t = 0.01\,s$) on a $64^2$ domain with $1\,mm$ spacing. Every time-series is an advection-diffusion process initialized by a Gaussian ($\mathcal{N}(0,\, 2)$) at a uniformly sampled center. The ``anomaly-free'' advection-diffusion parameters ($\overline{\mathbf{V}}$, $\overline{\mathbf{D}}$) are obtained from randomly generated potentials $\boldsymbol{\Psi},\, \mathbf{B},\, \boldsymbol{\Lambda}$ with uniformly sampled dominant diffusion directions. All components of $\boldsymbol{\Lambda}$ are randomly sampled from  $\mathcal{U}_{[0, 1]}$, and $\boldsymbol{\Psi}$ is randomly sampled from $\mathcal{U}_{[-10, 10]}$. %, to assure numerical stability of the simulation. 
Anomaly value fields are applied on the original ``anomaly-free'' parameters with a probability of 50\%, where the value ($\in[0, 1]$) and area of anomaly fields are determined by a multivariate Gaussian with its center and standard deviation uniformly sampled in the domain. The initial Gaussian is then transported by ${\mathbf{V}},\, {\mathbf{D}}$. (See Supp.~\ref{app: numericals} for details.)

%%%%%%%%%%%%%%%%%%%%%%%%%%%%
\vspace{-0.45cm}
\paragraph{Comparisons}
\label{sec: 2d_exp} 
To examine the contributions of different components in \texttt{D$^2$-SONATA}, we compare five variants (Tab.~\ref{tab: models}): (1) \texttt{YETI}: the original setting of \texttt{YETI} by Liu et al.~\cite{liu2021yeti}, where the model is trained with anisotropic Gaussians with no anomaly value field applied, and the advection-diffusion parameters are deterministically predicted without predicting an anomaly field  or uncertainty; (2) Anom.-\texttt{YETI}: here the \texttt{YETI} model remains unchanged, but is trained with anisotropic Gaussians of which 50\% are affected by the anomaly fields introduced in Sec.~\ref{sec: 2d_dataset}; (3) w/o-$\sigma$: \texttt{D$^2$-SONATA} without the stochastic term and the model uncertainty network $\mathcal{F}_{U}$; (4) w/o-$A$: \texttt{D$^2$-SONATA} without the decoder branch for the anomaly value field ($A$) in $\mathcal{F}_{P}$. In this case, the $\overline{\mathbf{V}}$, $\overline{\mathbf{D}}$-decoders directly output the overall value for $\mathbf{V}$, $\mathbf{D}$, instead of predicting the anomaly value field and corresponding anomaly-free parameters separately; (5) The full \texttt{D$^2$-SONATA} model.

We follow the training schedule in~\cite{liu2021yeti}, %, to compare fairly with \texttt{YETI}~\cite{liu2021yeti}. 
and re-train \texttt{YETI} with its original settings. The time-series sample length for all models is $N_{\text{in}} = 10$. We set $N_{\text{out}} = 10,\, w_{\mathbf{U}\boldsymbol{\Lambda}} = 0.5,\, w_{\text{SS}} = 0.1,\, w_{\sigma} = 0.5$. We test on 100 samples after every 500 training epochs and use the mean relative absolute error (RAE)~\cite{liu2021yeti} for the predicted $\widehat{\mathbf{V}}, \,\widehat{\mathbf{D}},\,\widehat{\mathbf{U}},\, \widehat{\boldsymbol{\Lambda}},\, \widehat{A}$:  
\vspace{-0.25cm}
\begin{equation}
Err(\mathbf{F}) = \frac{1}{|\Omega|}\int_{\Omega} \| \mathbf{F} - \widehat{\mathbf{F}} \|/\| \mathbf{F} \| \, d\mathbf{x}\,,
\label{eq: eval} 
\vspace{-0.35cm}
\end{equation}
where $\mathbf{F}(\widehat{\mathbf{F}})$ denotes the ground truth (prediction), $\|\cdot\|$ is the absolute, 2-norm and Frobenius norm for scalars, vectors and tensors. The time-series error ($Err(C)$) computes the RAE averaged over all predicted collocation time points. 

Fig.~\ref{fig: 2d_demo_loss_comp} shows the testing reconstruction errors of the five models throughout training. Being exposed to abnormal data (Anom.-\texttt{YETI}) during training significantly boosts \texttt{YETI}'s reconstruction performance. The anomaly value field ($A$) prediction helps $\mathbf{V},\, \mathbf{D}$'s reconstruction, and modeling as a stochastic system ($\sigma$) further improves the time-series predictions. Overall, with anomaly-prediction and stochastic modeling, \texttt{D$^2$-SONATA} outperforms all other variants with respect to both (ab-)normal time-series prediction and advection-diffusion parameter reconstruction.

%%%%%%%%%%%%%%%%%%%%%%%%%%%%%%%

%\input{sub/exp/fig/vd_generator} 

\vspace{-0.15cm}
\subsection{3D (Ab-)Normal Brain Advection-Diffusion}
\label{exp: ixi}
%%%%%%%%%%%%%%%%%%%%%%%%%%% 
\input{sub/exp/fig/ixi_plt}
\begin{table}[t]
\resizebox{\linewidth}{!}{
\centering
    %\begin{tabular}{P{0.6cm}P{1.3cm}|P{0.6cm}P{0.6cm}P{0.6cm}|P{0.6cm}P{0.6cm}|P{0.6cm}P{0.6cm}P{0.6cm}} 
    \begin{tabular}{ccccccc} 
       \toprule \\[-3ex] 
      % Models & \thead{\normalfont{Advection-diffusion}\\\normalfont{Parameters ($\mathbf{V,\, D}$)}} & \thead{\normalfont{``Anomaly-free''}\\\normalfont{Parameters ($\overline{\mathbf{V}},\, \overline{\mathbf{D}}$)}} & \thead{\normalfont{Anomaly Value}\\\normalfont{Field ($M$)}} & \thead{\normalfont{Uncertainty}\\\normalfont{($\sigma$)}} & Notes \\ [-0.2ex]
      {\bf{Models}} & $\widehat{C}$ & $\widehat{\mathbf{V}}$ & $\widehat{\mathbf{D}}$ & $\widehat{\mathbf{U}}$ & $\widehat{\boldsymbol{\Lambda}}$ & $\widehat{A}$  \\ [-0.2ex]
     \midrule\\[-3ex] 
     \multirow{1}{*}{\texttt{YETI}~\cite{liu2021yeti}} & 8.53 & 18.21 & 6.48 & 4.43 & 4.71 &  N/A   \\ 
     \hline\\[-2.4ex]
     \multirow{1}{*}{Anom.-\texttt{YETI}} & 2.52 & 12.23 & 5.15 & 4.12 & 3.54 &  N/A   \\ 
     \hline\\[-2.4ex]
     \multirow{1}{*}{w/o-$\sigma$} & 2.41 & 10.91 & 4.67 & 4.07 & 3.08 & 0.51   \\ 
     \hline\\[-2.4ex]
     \multirow{1}{*}{w/o-$A$} & 2.29 & 12.10 & 4.61 & 4.11 & 3.17 &  N/A   \\ 
     \hline\\[-2.4ex]
     \multirow{1}{*}{\texttt{D$^2$-SONATA}} & {\bf{2.12}} & {\bf{10.68}} & {\bf{4.21}} & {\bf{3.08}} & {\bf{2.94}} & {\bf{0.47}}    \\

\bottomrule  \\ [-3.3ex] 
	\vspace{-0.5cm}
    \end{tabular}  
    \caption{Reconstruction comparisons on mean RAE ($\%$) (Eq.~(\ref{eq: eval})) for 3D (ab-)normal brain advection-diffusion dataset (Sec.~\ref{exp: ixi}).} 
    \label{tab: ixi}
}

\end{table}
%\input{sub/exp/fig/ixi_loss}
%%%%%%%%%%%%%%%%%%%%%%%%%%%
\vspace{-0.1cm}
\paragraph{Dataset}
\label{sec: ixi_data}

We develop a normal-abnormal brain advection-diffusion simulator based on the \texttt{IXI} brain dataset\footnote{\url{http://brain-development.org/ixi-dataset/}}. Following the simulation procedure in~\cite{liu2021yeti}, we use 200 patients that have T1-/T2-weighted images, a magnetic resonance angiography (MRA) image, and diffusion-weighted images (DWI) with 15 directions, to simulate the normal 3D velocity and diffusion tensor fields\footnote{The goal is to obtain nontrivial advection-diffusion parameters for 3D simulation to boost a network's pre-training with supervision. This will likely not mimick real perfusion across the entire brain, but is expected to provide quasi-realistic local patterns which would appear in real data.}. For each case, we simulate both normal samples and anomaly-encoded samples, i.e., with simulated anomaly value fields applied to both $\overline{\mathbf{V}}$ and $\overline{\mathbf{D}}$, where the value (within $[0, 1]$) and area of the anomaly fields are calculated by a multivariate Gaussian with its center and standard deviation uniformly sampled over the spatial domain. Each resulting image has isotropic spacing ($1\,mm$) and has been rigidly registered intra-subject. In summary, we simulated 2,400 brain advection-diffusion time-series, where 50\% contain simulated anomalies. We randomly select 40 time-series for validation and testing respectively. (See Supp.~\ref{app: ixi} for more simulation details.)  

%%%%%%%%%%%%%%%%%%%%%%%
\vspace{-0.5cm}
\paragraph{Comparisons}
\label{sec: ixi_exp}

We compare the same five models as in Sec.~\ref{sec: 2d_exp} (Tab.~\ref{tab: models}). %, with input time-series length $N_{\text{in}} = 5$. 
We follow the training strategy in \texttt{YETI}~\cite{liu2021yeti} and re-train \texttt{YETI} with its original settings. Predicted parameters ($\widehat{\mathbf{V}},\, \widehat{\mathbf{D}},\, \widehat{\mathbf{U}},\, \widehat{\boldsymbol{\Lambda}}, \, \widehat{A}$) on the entire domain are obtained by splicing the output patches together. Integrating the stochastic advection-diffusion PDE forward with $\widehat{\mathbf{V}},\, \widehat{\mathbf{D}}$, we obtain the predicted time-series $\widehat{C}$ on the original domain. For better visualization, we use the $2$-norm ($\| \mathbf{V} \|_2$) map for the velocity fields, and the commonly used tensor feature maps~\cite{mukherjee2008dti,liu2021yeti} for the diffusion fields: (1) Trace ($tr_{\mathbf{D}}$), the sum of tensor eigenvalues ($\boldsymbol{\Lambda}$): the overall diffusion strength; and (2) Fractional anisotropy (FA): the amount of anisotropy across diffusing directions. %Mean RAE (Eq.~(\ref{eq: eval})) is used to evaluate the reconstruction performances on $\widehat{C}$, $\widehat{\mathbf{V}}$, and $tr_{\mathbf{D}}$, FA of $\widehat{\mathbf{D}}$.

%%%%%%%%%%%%%%%%%%%%%%%%%%%%%%%%%%%%%%%% 
Tab.~\ref{tab: ixi} compares the reconstruction errors of the predicted $\widehat{\mathbf{V}},\, \widehat{\mathbf{D}},\, \widehat{\mathbf{U}},\, \widehat{\boldsymbol{\Lambda}}$, and $\widehat{A}$ from the five models, where \texttt{D$^2$-SONATA} consistently outperforms {\it{all}} other models. Specifically, training with abnormal samples and the stochastic term ($\sigma$) to model uncertainties helps achieve better overall performance, particularly in predicting the transport time-series. Predicting with the proposed decomposition based on the anomaly value field ($A$) further boosts the reconstruction of the advection-diffusion parameters. 
The advantages of \texttt{D$^2$-SONATA} can be observed more clearly in Fig.~\ref{fig: ixi}, where it successfully captures the anomaly value field ($A$), and accurately reconstructs the overall magnitudes of the advection-diffusion parameters ($\mathbf{V},\, \mathbf{D}$). Without seeing abnormal samples during training, \texttt{YETI} struggles to identify the anomalies and to differentiate between advection ($\mathbf{V}$) and diffusion ($\mathbf{D}$) effectively ($2^{\text{nd}}$ column), where the resulting $tr_{\mathbf{D}}$ reflects insufficient differences between normal and abnormal regions and the reconstructed $\| \mathbf{V} \|_2$ is noisiest among all models. Training with abnormal samples generally helps the models to locate anomalies. Interestingly, decomposing into an anomaly value field and the ``anomaly-free'' parameters improves the reconstruction of the diffusion's fractional anisotropy (FA), especially within the abnormal regions. 

%%%%%%%%%%%%%%%%%%%%%%%%%%%%%%%
\vspace{-0.1cm}
\subsection{ISLES2017: Brain Perfusion Dataset from Ischemic Stroke Patients}
\label{exp: isles}

%%%%%%%%%%%%%%%%%%%%%%%
\vspace{-0.15cm}
\paragraph{Dataset}
\label{sec: isles_data}

We test on the Ischemic Stroke Lesion Segmentation (\texttt{ISLES}) 2017 dataset~\cite{isles2015a}, which contains perfusion data from 75 (43 training, 32 testing) ischemic stroke patients. Each patient has a dynamic susceptibility contrast (DSC) MR perfusion image (4D, with 40 to 80 available time points, time interval $\approx 1\,s$), along with the corresponding gold-standard lesion segmentation maps~\cite{essig2013mrp}. All images are resampled to isotropic spacing (1mm) and rigidly registered intra-subject via \texttt{ITK}~\cite{itk} following~\cite{liu2021yeti}. Perfusion images are converted to the tracer concentration time-series, $\{C^{t_i} \in \mathbb{R}(\Omega) \vert\, i = 1,\, 2,\, \ldots,\, N_T\}$ ($t_1$ is the time-to-peak for total concentration over the entire brain, at which the injected tracer is assumed to have been fully transported into the brain~\cite{liu2020piano,liu2021yeti}), via the relation between MR signal and tracer intensity~\cite{fieselmann2011sig2ctc}. 10 patients with lesion segmentation maps are randomly chosen for testing, while the remaining 65 concentration time-series are left-right flipped (i.e., the brain hemispheres are flipped) for data augmentation. In total, we obtain 130 time-series training samples, from which 10 samples are randomly picked for validation.

 %%%%%%%%%%%%%%%%%%%%
\input{sub/exp/fig/isles_plt}

\begin{table}[t]
\hspace*{-0.48cm} 
\resizebox{1.1\linewidth}{!}{
\centering 
    \begin{tabular}{P{0.5cm}P{0.6cm}P{0.75cm}P{0.75cm}P{0.75cm}P{0.75cm}P{0.75cm}P{0.75cm}P{0.75cm}P{0.75cm}P{0.75cm}P{0.75cm}}
    %\begin{tabular}{cccccccccccc}
       \toprule \\[-3ex] 
      \multicolumn{2}{c}{\multirow{2}{*}{\bf Metrics}} & \multicolumn{3}{c}{\bf \texttt{D$^2$-SONATA}} & \multicolumn{2}{c}{{\bf \texttt{YETI}}~\cite{liu2021yeti}} & \multicolumn{2}{c}{{\bf \texttt{PIANO}}~\cite{liu2020piano}}  &  \multicolumn{3}{c}{{\bf \texttt{ISLES}}~\cite{isles2015a}}   \\ [-0.7ex]
        \cmidrule(lr){3-5}% \\ [-4.1ex]
        \cmidrule(lr){6-7}%\\  
        \cmidrule(lr){8-9}%\\  
        \cmidrule(lr){10-12}%\\
       & & $A$ & ${\| {\bf{V}} \|_2}$ & $tr_{\mathbf{D}}$ & ${\| {\bf{V}} \|_2}$ & $tr_{\mathbf{D}}$ & ${\| {\bf{V}} \|_2}$ & $D$ & CBF & CBV & MTT \\ [-0.2ex]
     \midrule\\[-3ex]
     
      \multirow{3}{*}{\thead{$\mu^r$\\($\downarrow$)}} & {\emph{Me.}} & 0.47 & {\bf{0.29}} & 0.42 & 0.30 & 0.59 & 0.55 & 0.58 & 0.67 & 0.78 & 0.57 \\ [-0.4ex]
       &  {\emph{Med.}} & 0.49 & {\bf{0.30}} & 0.48 & 0.32 & 0.59 & 0.54 & 0.55 & 0.59 & 0.79 & 0.58 \\ %[0.7ex]
       &  {(\emph{STD})} & (0.13) & (0.17) & (0.15) & (0.11) & (0.19) & (0.15) & (0.16) & (0.12) & (0.23) & (0.13) \\ %[0.7ex]
     \hline\\[-2.3ex]
      \multirow{3}{*}{\thead{$|t|$\\($\uparrow$)}} & {\emph{Me.}} & {\bf 280} & 165 & 166 & 155 & 49 & 108 & 52 & 34 & 16 & 31  \\ %[0.7ex]
       &  {\emph{Med.}} & {\bf 286} & 164 & 158 & 134 & 42 & 89 & 48 & 28 & 11 & 32 \\ %[0.7ex]
       &  {(\emph{STD})} & (58) & (37) & (60) & (62) & (22) & (35) & (26) & (22) & (12) & (37)  \\ %[0.7ex] % [2ex] 
     \hline\\[-2.4ex]
      \multirow{3}{*}{\thead{\normalfont{AUC}\\($\uparrow$)}} &  {\emph{Me.}} & {\bf 0.79} & 0.70 & 0.64 & 0.73 & 0.51 & 0.74 & 0.68 & 0.72 & 0.65 & 0.65  \\ %[0.7ex]
       &  {\emph{Med.}} & {\bf 0.76} & 0.71 & 0.65 & 0.73 & 0.50 & 0.74 & 0.69 & 0.73 & 0.68 & 0.66 \\ %[0.7ex]
       &  {(\emph{STD})} & (0.05) & (0.04) & (0.07) & (0.06) & (0.03) & (0.04) & (0.03) & (0.07) & (0.06) & (0.06) \\ [-0.5ex] 
      
\midrule[\heavyrulewidth] \\ [-3.5ex]
\multicolumn{8}{l}{\footnotesize$^*$ $\downarrow$ ($\uparrow$) indicates the lower (higher) values are better.} \\ [-0.5ex]
\bottomrule  \\ [-3.6ex] 
    \end{tabular}  
	\vspace{-1cm}
    \caption{Quantitative comparison between \texttt{D$^2$-SONATA}, \texttt{YETI}, \texttt{PIANO} and \texttt{ISLES} maps across 10 test subjects from \texttt{ISLES2017} dataset (Sec.~\ref{sec: isles_exp}), using \emph{Mean (Me.)}, \emph{Median (Med.)}, \emph{Standard Deviation (STD)} of relative mean $\mu^r$, absolute ($|t|$), and area under the curve (AUC).} 
    \label{tab: isles}
}

\end{table}

 %%%%%%%%%%%%%%%%%%%%
 
%%%%%%%%%%%%%%%%%%%%%%% 
\vspace{-0.5cm}
\paragraph{Results}
\label{sec: isles_exp}

We transfer the pre-trained model from our simulated 3D normal-abnormal brain advection-diffusion dataset (Sec.~\ref{sec: ixi_exp}) to the \texttt{ISLES} tracer concentration time-series dataset, using the \texttt{Adam} optimizer and learning rate $10^{-4}$. We set $N_{\text{in}} = N_{\text{out}} = 5$, $w_{\sigma} = 0.5,\, w_{\text{SS}} = 0.1$.

Following the feature maps for comparison proposed in~\cite{liu2020piano,liu2021yeti}, we compute (1) the $2$-norm ($\|\mathbf{V}\|_2$) for the velocity field, and (2) the trace ($tr_{\mathbf{D}}$) for diffusion tensors (both introduced in Sec.~\ref{sec: ixi_exp}). In \texttt{D$^2$-SONATA}, we also show (3) the output model uncertainty ($\sigma$), (4) the predicted anomaly value field ($A$), and (5) $Seg(A)$, the segmentation map obtained by thresholding\footnote{The chosen threshold is the optimal threshold for the segmentation Dice score, averaged from 6 randomly-selected test patients.} $A$.

Fig.~\ref{fig: isles_plt} shows the feature maps for four test patients from our model. All show features consistent with the manually segmented lesion maps. Importantly, besides the feature maps of the predicted advection-diffusion parameters ($\| \mathbf{V}\|,\, tr_{\mathbf{D}}$), we additionally obtain (1) the anomaly value field ($A$) from which one can directly obtain the lesion segmentation ($Seg(A)$), and (2) the ``anomaly-free'' parameters ($\| \overline{\mathbf{V}}\|,\, tr_{\overline{\mathbf{D}}}$) which provide insights into the expected normal brain perfusion of a patient.
 
%%%%%%%%%%%%%%%%%%%%%%%
\vspace{-0.45cm}
\paragraph{Comparisons} 
We compare the performance of feature maps from \texttt{D$^2$-SONATA} ($\|\mathbf{V}\|_2,\, \mathbf{D},\, A$) with (1) \texttt{YETI}~\cite{liu2021yeti}: $\|\mathbf{V}\|_2,\, \mathbf{D}$; (2) \texttt{PIANO}~\cite{liu2020piano}: $\|\mathbf{V}\|_2,\, D$\footnote{We directly use $D$ for \texttt{PIANO} as it models diffusion as scalar fields.}; and (2) \texttt{ISLES}~\cite{isles2015a} (perfusion summary maps): cerebral blood flow (CBF), cerebral blood volume (CBV), mean transit time (MTT). We report the two measures proposed by Liu et al.~\cite{liu2021yeti} that focus on the feature map differences between lesions and normal regions:
(1) Relative mean ($\mu^r \in [0, 1]$):
\vspace{-0.25cm}  
\begin{equation} 
\mu^r = min\bigg\{\frac{\text{mean in lesion}}{\text{mean in c-lesion}}, \frac{\text{mean in c-lesion}}{\text{mean in lesion}}\bigg\},
\label{eq: rel_mean} 
\vspace{-0.2cm}
\end{equation} 
where c-lesion refers to the contralateral region of a lesion obtained by mirroring via the midline of the cerebral hemispheres, and $min$ accounts for typically larger MTT values (while other metrics are typically smaller) in the lesion rather than the c-lesion region; (2) Absolute t-value ($|t|$): the absolute value of the unpaired t-statistic between voxels in the lesion and the c-lesion regions.

The above two measures are defined given a known segmented lesion region. In this work, we additionally consider how well the feature maps can distinguish lesion from non-lesion regions. I.e., we want to directly use them for lesion segmentation. To this end we threshold all feature maps and additionally compare the resulting areas under the curve (AUC) of the receiver operating characteristic (ROC) curves computed from all feature maps.

Tab.~\ref{tab: isles} compares feature maps from \texttt{D$^2$-SONATA}, \texttt{YETI}, \texttt{PIANO}, and \texttt{ISLES} based on the above metrics for the 10 test patients. Despite predicting the advection-diffusion parameters in the form of an anomaly-encoded decomposition, \texttt{D$^2$-SONATA} still results in  the lowest relative means ($\mu^r$) between lesion and c-lesion for $\| \mathbf{V} \|_2$ and $tr_{\mathbf{D}}$. This  indicates \texttt{D$^2$-SONATA}'s stronger ability to differentiate between normal and abnormal fields. Moreover, the predicted anomaly value field ($A$) obtains a much higher absolute t-value ($|t|$) which indicates great potential for lesion segmentation. This is confirmed  by $A$ achieving the highest AUC score among all feature maps.

\section{Conclusions}
\label{sec: con}
\vspace{-0.15cm}
We presented \texttt{D$^2$-SONATA}, a two-stage learning scheme to estimate the velocity vector fields and diffusion tensor fields that drive real-world-observed advection-diffusion processes in the presence of anomalies. The model is built upon a stochastic advection-diffusion PDE and predicts divergence-free velocity fields and symmetric PSD diffusion tensor fields by construction. Further, our model directly predicts an anomaly field and provides measures of uncertainty. Extensive experiments on 2D and 3D simulated data illustrate that our model can successfully reconstruct the advection-diffusion fields (with or without anomalies) from the time-series data of observed transport processes. When applied to real brain perfusion data from ischemic stroke patients, our model outperforms the state of the art across {\it{all}} metrics. By predicting the underlying advection-diffusion parameters along with an anomaly value field our model provides additional insights into real stroke data. 

%\section*{Acknowledgment}
%Research reported in this work was supported by the National Institutes of Health (NIH) under awards number NIH 2R42NS086295 and NIH R21NS125369. The content is solely the responsibility of the authors and does not necessarily represent the official views of the NIH.

%\subsubsection*{Acknowledgments}

%%%%%%%%%%%%%%%%%%%%%%%%%

%\newpage
{\small
\bibliographystyle{ieee_fullname}
\bibliography{CVPR22-229}

\begin{thebibliography}{10}\itemsep=-1pt

\bibitem{2dunet2015}
{{U}-{N}et: Convolutional Networks for Biomedical Image Segmentation},
  author={Olaf Ronneberger and Philipp Fischer and Thomas Brox}, year={2015},
  journal={arXiv preprint arXiv:1505.04597},.

\bibitem{amrouche1998div_free}
Ch{\'{e}}rif Amrouche, Christine Bernardi, Monique Dauge, and Vivette Girault.
\newblock Vector potentials in three-dimensional non-smooth domains.
\newblock {\em Mathematical Methods in the Applied Sciences}, 21(9):823--864,
  1998.

\bibitem{amrouche2013div_free}
Ch{\'{e}}rif Amrouche and Nour El~Houda Seloula.
\newblock Lp-theory for vector potentials and {S}obolev's inequalities for
  vector fields: Application to the {S}tokes equations with pressure boundary
  conditions.
\newblock {\em Mathematical Models and Methods in Applied Sciences},
  23(01):37--92, 2013.

\bibitem{Guerra2009brownian}
Jo\ ao Guerra and David Nualart.
\newblock Stochastic differential equations driven by fractional {B}rownian
  motion and standard brownian motion.
\newblock {\em Stochastic Analysis and Applications}, 26(5):1053--1075, 2008.

\bibitem{balakrishnan2019voxelmorph}
Guha Balakrishnan, Amy Zhao, Mert~R Sabuncu, John Guttag, and Adrian~V Dalca.
\newblock Voxelmorph: a learning framework for deformable medical image
  registration.
\newblock {\em IEEE transactions on medical imaging}, 38(8):1788--1800, 2019.

\bibitem{bar2019pde}
Yohai Bar-Sinai, Stephan Hoyer, Jason Hickey, and Michael~P. Brenner.
\newblock Learning data-driven discretizations for partial differential
  equations.
\newblock {\em Proceedings of the National Academy of Sciences},
  116(31):15344--15349, 2019.

\bibitem{beg2005computing}
Mirza~Faisal {Beg}, Michael~I. {Miller}, Alain Trouv{\'e}, and Laurent
  {Younes}.
\newblock Computing large deformation metric mappings via geodesic flows of
  diffeomorphisms.
\newblock {\em International journal of computer vision}, 61(2):139--157, 2005.

\bibitem{bhattacharya2020reduction}
Kaushik Bhattacharya, Bamdad Hosseini, Nikola~B. Kovachki, and Andrew~M.
  Stuart.
\newblock Model reduction and neural networks for parametric {PDE}s.
\newblock {\em arXiv preprint arXiv:2005.03180}, 2020.

\bibitem{borzi2003optimal}
Alfio Borzi, Kazufumi Ito, and Karl Kunisch.
\newblock Optimal control formulation for determining optical flow.
\newblock {\em SIAM journal on scientific computing}, 24(3):818--847, 2003.

\bibitem{3dunet2016}
{\"O}zg{\"u}n {\c{C}}i{\c{c}}ek, Ahmed Abdulkadir, Soeren~S. Lienkamp, Thomas
  Brox, and Olaf Ronneberger.
\newblock 3{D} {U}-{N}et: Learning dense volumetric segmentation from sparse
  annotation.
\newblock In Sebastien Ourselin, Leo Joskowicz, Mert~R. Sabuncu, Gozde Unal,
  and William Wells, editors, {\em Medical Image Computing and
  Computer-Assisted Intervention -- MICCAI 2016}, pages 424--432, Cham, 2016.
  Springer International Publishing.

\bibitem{cookson2014spatial}
Andrew Cookson, Jack Lee, Christian Michler, Radomir Chabiniok, Eoin~R Hyde,
  David Nordsletten, and Nicolas Smith.
\newblock A spatially-distributed computational model to quantify behaviour of
  contrast agents in {MR} perfusion imaging.
\newblock {\em Medical Image Analysis}, 18(7):1200--1216, 2014.

\bibitem{emmanuel2018}
Emmanuel de B{\'{e}}zenac, Arthur Pajot, and Patrick Gallinari.
\newblock Deep learning for physical processes: Incorporating prior scientific
  knowledge.
\newblock In {\em 6th International Conference on Learning Representations,
  {ICLR} 2018, Vancouver, BC, Canada, April 30 - May 3, 2018, Conference Track
  Proceedings}, 2018.

\bibitem{demeestere2020stroke}
Jelle Demeestere, Anke Wouters, Soren Christensen, Robin Lemmens, and
  Maarten~G. Lansberg.
\newblock Review of perfusion imaging in acute ischemic stroke.
\newblock {\em Stroke}, 51(3):1017--1024, 2020.

\bibitem{flownet}
Alexey {Dosovitskiy}, Philipp {Fischer}, Eddy Ilg, Philip {H\"ausser}, Caner
  {Hazirbas}, Vladimir {Golkov}, Patrick van~der {Smagt}, Daniel {Cremers}, and
  Thomas {Brox}.
\newblock Flownet: Learning optical flow with convolutional networks.
\newblock In {\em 2015 IEEE International Conference on Computer Vision
  (ICCV)}, pages 2758--2766, 2015.

\bibitem{dubois1990div_free}
Francois Dubois.
\newblock Discrete vector potential representation of a divergence-free vector
  field in three-dimensional domains: Numerical analysis of a model problem.
\newblock {\em SIAM Journal on Numerical Analysis}, 27(5):1103--1141, 1990.

\bibitem{weinan2018ritz}
Weinan E and Bing Yu.
\newblock The deep {R}itz method: A deep learning-based numerical algorithm for
  solving variational problems.
\newblock {\em Communications in Mathematics and Statistics},
  6(1):e2019WR026731, 2018.

\bibitem{essig2013mrp}
Marco Essig, Mark~S Shiroishi, Thanh~Binh Nguyen, Marc Saake, James~M
  Provenzale, David Enterline, Nicoletta Anzalone, Arnd D\"orfler, Alex Rovira,
  Max Wintermark, and Meng Law.
\newblock Perfusion {MRI}: the five most frequently asked technical questions.
\newblock {\em AJR. American journal of roentgenology}, 200(1):24--34, 2013.

\bibitem{evans2000pde}
Gwynne~A. Evans, Jonathan~M. Blackledge, and Peter~D. Yardley.
\newblock {\em Numerical Methods for Partial Differential Equations}.
\newblock Springer, 2000.

\bibitem{fieselmann2011sig2ctc}
Andreas Fieselmann, Markus Kowarschik, Arundhuti Ganguly, Joachim Hornegger,
  and Rebecca Fahrig.
\newblock Deconvolution-based {CT} and {MR} brain perfusion measurement:
  Theoretical model revisited and practical implementation details.
\newblock {\em Journal of Biomedical Imaging}, 2011, 2011.

\bibitem{gottlieb2000ssp}
Sigal Gottlieb and Lee-Ad~J. Gottlieb.
\newblock Strong stability preserving properties of {Runge-Kutta} time
  discretization methods for linear constant coefficient operators.
\newblock {\em Journal of Scientific Computing}, pages 83--109, 2003.

\bibitem{pet2011}
Julie~Marie Gr\"u~ner, Rune Paamand, Liselotte H\o~jgaard, and Ian Law.
\newblock Brain perfusion {CT} compared with
  $^{\text{15}}${O}-$\text{H}_{\text{2}}${O}-{PET} in healthy subjects.
\newblock {\em EJNMMI research}, 1(1), 2011.

\bibitem{harabis2013dilution}
Vratislav Harabis, Radim Kolar, Martin Mezl, and Radovan Jirik.
\newblock Comparison and evaluation of indicator dilution models for bolus of
  ultrasound contrast agents.
\newblock {\em Physiological measurement}, 34(2):151--162, 2013.

\bibitem{hart2009optimal}
Gabriel~L. Hart, Christopher Zach, and Marc Niethammer.
\newblock An optimal control approach for deformable registration.
\newblock In {\em 2009 IEEE Computer Society Conference on Computer Vision and
  Pattern Recognition (CVPR) Workshops}, pages 9--16. IEEE, 2009.

\bibitem{horn1981determining}
Berthold~K.P. Horn and Brian~G. Rhunck.
\newblock Determining optical flow.
\newblock In {\em Techniques and Applications of Image Understanding}, volume
  281, pages 319--331. International Society for Optics and Photonics, 1981.

\bibitem{horn1980optflow}
Berthold~K.P. Horn and Brian~G. Schunck.
\newblock Determining optical flow.
\newblock Technical report, 1980.

\bibitem{kim19deepfluid}
Byungsoo Kim, Vinicius C.~Azevedo, Nils Thuerey, Theodore Kim, Markus Gross,
  and Barbara Solenthaler.
\newblock Deep fluids: A generative network for parameterized fluid
  simulations.
\newblock {\em Computer Graphics Forum (Proc. Eurographics)}, 38(2), 2019.

\bibitem{Kloeden1992Stochastic}
Peter~E. Kloeden and Eckhard Platen.
\newblock {\em Numerical Solution of Stochastic Differential Equations}.
\newblock Springer, 2000.

\bibitem{kong2020sde}
Lingkai Kong, Jimeng Sun, and Chao Zhang.
\newblock {SDE}-{N}et: Equipping deep neural networks with uncertainty
  estimates.
\newblock In {\em International Conference on Machine Learning (ICML)}, 2020.

\bibitem{koundal2020omt}
Sunil Koundal, Rena Elkin, Saad Nadeem, Yuechuan Xue, Stefan Constantinou,
  Simon Sanggaard, Xiaodan Liu, Brittany Monte, Feng Xu, William Van~Nostrand,
  Maiken Nedergaard, Hedok Lee, Joanna Wardlaw, Helene Benveniste, and Allen
  Tannenbaum.
\newblock Optimal mass transport with {L}agrangian workflow reveals advective
  and diffusion driven solute transport in the glymphatic system.
\newblock {\em Scientific Reports}, 10(1):1990, 2020.

\bibitem{leveque2002}
Randall~J. LeVeque.
\newblock {\em Finite Volume Methods for Hyperbolic Problems}.
\newblock Cambridge Texts in Applied Mathematics. Cambridge University Press,
  2002.

\bibitem{lezcano2019trivializations}
Mario Lezcano-Casado.
\newblock Trivializations for gradient-based optimization on manifolds.
\newblock In {\em Advances in Neural Information Processing Systems (NeurIPS)},
  pages 9154--9164, 2019.

\bibitem{lezcano2019spectral}
Mario Lezcano-Casado and David Mart{\'i}nez-Rubio.
\newblock Cheap orthogonal constraints in neural networks: A simple
  parametrization of the orthogonal and unitary group.
\newblock In {\em International Conference on Machine Learning (ICML)}, pages
  3794--3803, 2019.

\bibitem{li2020fourier}
Zongyi Li, Nikola Kovachki, Kamyar Azizzadenesheli, Burigede Liu, Kaushik
  Bhattacharya, Andrew Stuart, and Anima Anandkumar.
\newblock Fourier neural operator for parametric partial differential
  equations.
\newblock {\em arXiv preprint arXiv:2010.08895}, 2020.

\bibitem{lieberman2010invpde}
Chad Lieberman, Karen Willcox, and Omar Ghattas.
\newblock Parameter and state model reduction for large-scale statistical
  inverse problems.
\newblock {\em SIAM J. Sci. Comput.}, 32(5):2523--2542, Aug. 2010.

\bibitem{liu2020piano}
Peirong Liu, Yueh~Z. Lee, Stephen~R. Aylward, and Marc Niethammer.
\newblock {PIANO}: Perfusion imaging via advection-diffusion.
\newblock In {\em Medical Image Computing and Computer Assisted Intervention
  (MICCAI)}, 2020.

\bibitem{liu2021tmi}
Peirong Liu, Yueh~Z. Lee, Stephen~R. Aylward, and Marc Niethammer.
\newblock Perfusion imaging: An advection diffusion approach.
\newblock {\em IEEE Transactions on Medical Imaging}, 2021.

\bibitem{liu2021yeti}
Peirong Liu, Lin Tian, Yubo Zhang, Stephen Aylward, Yueh Lee, and Marc
  Niethammer.
\newblock Discovering hidden physics behind transport dynamics.
\newblock In {\em Proceedings of the IEEE/CVF Conference on Computer Vision and
  Pattern Recognition (CVPR)}, 2021.

\bibitem{lu2020deeponet}
Lu Lu, Pengzhan Jin, and George~Em Karniadakis.
\newblock Deeponet: Learning nonlinear operators for identifying differential
  equations based on the universal approximation theorem of operators.
\newblock {\em arXiv preprint arXiv:1910.03193}, 2020.

\bibitem{isles2015a}
Oskar Maier, Bjoern~H Menze, Janina von~der Gablentz, Levin Hani, Mattias~P
  Heinrich, Matthias Liebrand, Stefan Winzeck, Abdul Basit, Paul Bentley, Liang
  Chen, Daan Christiaens, Francis Dutil, Karl Egger, Chaolu Feng, Ben Glocker,
  Michael G\"otz, Tom Haeck, Hanna-Leena Halme, Mohammad Havaei, Khan~M
  Iftekharuddin, Pierre-Marc Jodoin, Konstantinos Kamnitsas, Elias Kellner,
  Antti Korvenoja, Hugo Larochelle, Christian Ledig, Jia-Hong Lee, Frederik
  Maes, Qaiser Mahmood, Klaus~H Maier-Hein, Richard McKinley, John Muschelli,
  Chris Pal, Linmin Pei, Janaki~Raman Rangarajan, Syed M~S Reza, David Robben,
  Daniel Rueckert, Eero Salli, Paul Suetens, Ching-Wei Wang, Matthias Wilms,
  Jan~S Kirschke, Ulrike~M Kr\"amer, Thomas~F M\"unte, Peter Schramm, Roland
  Wiest, Heinz Handels, and Mauricio Reyes.
\newblock {ISLES} 2015 - a public evaluation benchmark for ischemic stroke
  lesion segmentation from multispectral {MRI} medical image analysis.
\newblock {\em Medical Image Analysis}, 35, 2017.

\bibitem{maria2003hhd}
Filippo Maria~Denaro.
\newblock On the application of the {H}elmholtz-{H}odge decomposition in
  projection methods for incompressible flows with general boundary conditions.
\newblock {\em International Journal for Numerical Methods in Fluids},
  43(1):43--69, 2003.

\bibitem{mathieu2016gdl}
Micha{\"{e}}l Mathieu, Camille Couprie, and Yann LeCun.
\newblock Deep multi-scale video prediction beyond mean square error.
\newblock In {\em 4th International Conference on Learning Representations,
  {ICLR} 2016, San Juan, Puerto Rico, May 2-4, 2016, Conference Track
  Proceedings}, 2016.

\bibitem{itk}
Matthew McCormick, Xiaoxiao Liu, Julien Jomier, Charles Marion, and Luis
  Ibanez.
\newblock {ITK}: enabling reproducible research and open science.
\newblock {\em Frontiers in Neuroinformatics}, 8(13), 2014.

\bibitem{modersitzki2004_numerical}
Jan Modersitzki.
\newblock {\em Numerical methods for image registration}.
\newblock Oxford University Press on Demand, 2004.

\bibitem{mouridsen2006aif}
Kim Mouridsen, Søren Christensen, Louise Gyldensted, and Leif {\O}stergaard.
\newblock Automatic selection of arterial input function using cluster
  analysis.
\newblock {\em Magnetic Resonance in Medicine}, 55(3):524--531, 2006.

\bibitem{mukherjee2008dti}
Pratik Mukherjee, Jeffrey Berman, Stephen~W. Chung, Christopher Hess, and
  Roland Henry.
\newblock Diffusion tensor {MR} imaging and fiber tractography: Theoretic
  underpinnings.
\newblock {\em American Journal of Neuroradiology}, 29(4):632--641, 2008.

\bibitem{nelsen2020random}
Nicholas~H. Nelsen and Andrew~M. Stuart.
\newblock The random feature model for input-output maps between banach spaces.
\newblock {\em arXiv preprint arXiv:2005.10224}, 2020.

\bibitem{niethammer2006dti}
Marc {Niethammer}, Raul San~Jose {Estepar}, Sylvain {Bouix}, Martha {Shenton},
  and Carl-Fredrik {Westin}.
\newblock On diffusion tensor estimation.
\newblock In {\em 2006 International Conference of the IEEE Engineering in
  Medicine and Biology Society}, pages 2622--2625, 2006.

\bibitem{niethammer2009optimal}
Marc Niethammer, Gabriel~L Hart, and Christopher Zach.
\newblock An optimal control approach for the registration of image
  time-series.
\newblock In {\em Proceedings of the 48h IEEE Conference on Decision and
  Control (CDC) held jointly with 2009 28th Chinese Control Conference}, pages
  2427--2434. IEEE, 2009.

\bibitem{oksendal2010stochastic}
B. {\O}ksendal.
\newblock {\em Stochastic Differential Equations: An Introduction with
  Applications}.
\newblock Universitext. Springer Berlin Heidelberg, 2010.

\bibitem{asl2010}
Sasitorn Petcharunpaisan, Joana Ramalho, and Mauricio Castillo.
\newblock Arterial spin labeling in neuroimaging.
\newblock {\em World journal of radiology}, 2(10):384--398, 2010.

\bibitem{pinn2019}
Maziar {Raissi}, Paris {Perdikaris}, and George~Em {Karniadakis}.
\newblock Physics-informed neural networks: A deep learning framework for
  solving forward and inverse problems involving nonlinear partial differential
  equations.
\newblock {\em Journal of Computational Physics}, 378:686 -- 707, 2019.

\bibitem{schmainda2019a}
K.M. Schmainda, M.A. Prah, L.S. Hu, C.C. Quarles, N. Semmineh, S.D. Rand, J.M.
  Connelly, B. Anderies, Y. Zhou, Y. Liu, B. Logan, A. Stokes, G. Baird, and
  J.L. Boxerman.
\newblock Moving toward a consensus {DSC-MRI} protocol: Validation of a
  low{\textendash}flip angle single-dose option as a reference standard for
  brain tumors.
\newblock {\em American Journal of Neuroradiology}, 2019.

\bibitem{schmainda2019b}
K.M. Schmainda, M.A. Prah, Z. Zhang, B.S. Snyder, S.D. Rand, T.R. Jensen, D.P.
  Barboriak, and J.L. Boxerman.
\newblock Quantitative delta {T}1 (d{T}1) as a replacement for adjudicated
  central reader analysis of contrast-enhancing tumor burden: A subanalysis of
  the american college of radiology imaging network 6677/radiation therapy
  oncology group 0625 multicenter brain tumor trial.
\newblock {\em American Journal of Neuroradiology}, 2019.

\bibitem{shen2019networks}
Zhengyang Shen, Xu Han, Zhenlin Xu, and Marc Niethammer.
\newblock Networks for joint affine and non-parametric image registration.
\newblock In {\em Proceedings of the IEEE Conference on Computer Vision and
  Pattern Recognition (CVPR)}, pages 4224--4233, 2019.

\bibitem{sitzmann2020implicit}
Vincent Sitzmann, Julien N.~P. Martel, Alexander~W. Bergman, David~B. Lindell,
  and Gordon Wetzstein.
\newblock Implicit neural representations with periodic activation functions.
\newblock {\em arXiv preprint arXiv:2006.09661}, 2020.

\bibitem{smith2020eikonet}
Jonathan~D. Smith, Kamyar Azizzadenesheli, and Zachary~E. Ross.
\newblock Eikonet: Solving the eikonal equation with deep neural networks.
\newblock {\em arXiv preprint arXiv:2004.00361}, 2020.

\bibitem{strouthos2010dilution}
Costas Strouthos, Marios Lampaskis, Vassilis Sboros, Alan Mcneilly, and
  Michalakis Averkiou.
\newblock Indicator dilution models for the quantification of microvascular
  blood flow with bolus administration of ultrasound contrast agents.
\newblock {\em IEEE Transactions on Ultrasonics, Ferroelectrics, and Frequency
  Control}, 57(6):1296--1310, 2010.

\bibitem{sun2010optflow}
Deqing Sun, Stefan Roth, and Michael~J. Black.
\newblock Secrets of optical flow estimation and their principles.
\newblock In {\em Proceedings of the IEEE/CVF Conference on Computer Vision and
  Pattern Recognition (CVPR)}, pages 2432--2439, 2010.

\bibitem{tartakovsky2020pinn}
Alexandre~M. {Tartakovsky}, Carlos~Ortiz {Marrero}, Paris {Perdikaris},
  Guzel~D. {Tartakovsky}, and David {Barajas-Solano}.
\newblock Physics-informed deep neural networks for learning parameters and
  constitutive relationships in subsurface flow problems.
\newblock {\em Water Resources Research}, 56(5):e2019WR026731, 2020.

\bibitem{Demian2014probreg}
Demian Wassermann, Matthew Toews, Marc Niethammer, and William Wells.
\newblock Probabilistic diffeomorphic registration: Representing uncertainty.
\newblock In S{\'e}bastien Ourselin and Marc Modat, editors, {\em Biomedical
  Image Registration}, pages 72--82, Cham, 2014. Springer International
  Publishing.

\bibitem{deepflow2013}
Philippe {Weinzaepfel}, J\'er\^ome {Revaud}, Zaid {Harchaoui}, and Cordelia
  {Schmid}.
\newblock Deepflow: Large displacement optical flow with deep matching.
\newblock In {\em 2013 IEEE International Conference on Computer Vision}, pages
  1385--1392, 2013.

\bibitem{yang2017quicksilver}
Xiao Yang, Roland Kwitt, Martin Styner, and Marc Niethammer.
\newblock Quicksilver: Fast predictive image registration--a deep learning
  approach.
\newblock {\em NeuroImage}, 158:378--396, 2017.

\bibitem{zhang2021qtm}
Qihao Zhang, Pascal Spincemaille, Michele Drotman, Christine Chen, Sarah
  Eskreis-Winkler, Weiyuan Huang, Liangdong Zhou, John Morgan, Thanh~D. Nguyen,
  Martin~R. Prince, and Yi Wang.
\newblock Quantitative transport mapping ({QTM}) for differentiating benign and
  malignant breast lesion: Comparison with traditional kinetics modeling and
  semi-quantitative enhancement curve characteristics.
\newblock {\em Magnetic Resonance Imaging}, 2021.

\bibitem{zhou2018vector}
Liangdong Zhou, Pascal Spincemaille, Qihao Zhang, Thanh~D. Nguyen, and Yi Wang.
\newblock Vector field perfusion imaging: A validation study by using
  multiphysics model.
\newblock {\em Proc. Intl. Soc. Mag. Recon. Med. (ISMRM)}, 26, 2018.

\bibitem{zhou2021qtm}
Liangdong Zhou, Qihao Zhang, Pascal Spincemaille, Thanh~D. Nguyen, John Morgan,
  Weiying Dai, Yi Li, Ajay Gupta, Martin~R. Prince, and Yi Wang.
\newblock Quantitative transport mapping ({QTM}) of the kidney with an
  approximate microvascular network.
\newblock {\em Magnetic Resonance in Medicine}, 85(4):2247--2262, 2021.

\end{thebibliography}
}

%%%%%%%%%%%%%%%%%%%%%%%%%

\clearpage
\onecolumn
\appendix

\renewcommand\thefigure{\thesection.\arabic{figure}}

{\centering 
\section*{Deep Decomposition for Stochastic Normal-Abnormal Transport\\Supplementary Material}
}
\vspace{0.55cm}
This supplementary material contains proofs for our representation theorems and additional implementation details for \texttt{D$^2$-SONATA}. Supp.~\ref{app: proof_div_free} and Supp.~\ref{app: proof_psd} give the proofs for Theorem~\ref{thm: repre_v} and Theorem~\ref{thm: repre_psd}, respectively. Supp.~\ref{app: numericals} introduces our 2D/3D stochastic advection-diffusion PDE package in \texttt{PyTorch} and discusses numerical discretization and stability conditions. And Supp.~\ref{app: ixi} provides descriptions of how we generate our normal-abnormal brain advection-diffusion simulation dataset, including how we construct the velocity and diffusion fields with random abnormal patterns, and how we simulate the corresponding normal and abnormal brain advection-diffusion time-series.

\vspace{0.5cm}

%%%%%%%%%%%%%%%%%%%%%%%%%%%%%%%

%\setcounter{page}{1} % for submission only, remove for arxiv version.

%%%%%%%%%%%%%%%%%%%%%%%%%%%%%%%

\section{Theorem \ref{thm: repre_v}: Anomaly-decomposed Divergence-free Vector Representation}
\label{app: proof_div_free}

{\it
\noindent For any vector field $\mathbf{V} \in L^p(\Omega)^d$ and scalar field $A$ in $\mathbb{R}_{(0,\, 1]}(\Omega)$ on a bounded domain $\Omega\subset\mathbb{R}^d$ with smooth boundary $\partial \Omega$. If $\mathbf{V}$ satisfies $\nabla \cdot \mathbf{V} = 0$, there exist a potential $\boldsymbol{\Psi}$ in $L^p(\Omega)^{\alpha}$ ($\alpha = 1(3)$ when $d = 2(3)$):
\begin{equation}
\mathbf{V} = \nabla \times (A \, \boldsymbol{\Psi}), \quad (A \, \boldsymbol{\Psi}) \cdot \mathbf{n}\big|_{\partial \Omega} = 0.
\label{app_eq: repre_v} 
\end{equation}
Conversely, for any $A \in \mathbb{R}_{(0,\, 1]}(\Omega)$, $\boldsymbol{\Psi}\in L^p(\Omega)^{\alpha}$, $\nabla \cdot \mathbf{V} =\nabla \cdot (\nabla \times (A \, \boldsymbol{\Psi}))= 0$.  
}

{\noindent  (Here, $L^{p}$ refers to the space of measurable functions for which the $p$-th power of the function absolute value is Lebesgue integrable. Specifically, let $1 \leq p < \infty$ and $(\Omega, \Sigma, \mu)$ be a measure space. $L^p(\Omega)$ space is the set of all measurable functions whose absolute value raised to the $p$-th power has a finite integral, i.e.,
$\|f\|_p \equiv \left( \int_{\Omega} |f|^p\;\mathrm{d}\mu \right)^{1/p}<\infty$.)}

\begin{proof}
The above Theorem is a Corollary of the following Theorem \ref{app_thm: repre_div_free}, which is originally introduced by Liu et al. \cite{liu2021yeti}. 

\begin{app_theorem}[Divergence-free Vector Field Representation by the Curl of Potentials]For any vector field $\mathbf{V} \in L^p(\Omega)^d$ on a bounded domain $\Omega\subset\mathbb{R}^d$ with smooth boundary $\partial \Omega$. If $\mathbf{V}$ satisfies $\nabla \cdot \mathbf{V} = 0$, there exists a potential $\boldsymbol{\Psi}$ in $L^p(\Omega)^{\alpha}$ such that ($\alpha = 1$ when $d = 2$, $\alpha = 3$ when $d = 3$)
\begin{equation}
\mathbf{V} = \nabla \times \boldsymbol{\Psi}, \quad \boldsymbol{\Psi} \cdot \mathbf{n}\big|_{\partial \Omega} = 0,\, \boldsymbol{\Psi} \in L^{p}(\Omega)^{\alpha}.
%\label{eq: repre_v}
\end{equation}
Conversely, for any $\boldsymbol{\Psi}\in L^p(\Omega)^{\alpha}$, $\nabla \cdot \mathbf{V} =\nabla \cdot (\nabla \times \boldsymbol{\Psi})= 0$. 
\label{app_thm: repre_div_free}

\noindent (For detailed proof of Theorem \ref{app_thm: repre_div_free}, please refer to Supp.~A in \cite{liu2021yeti}.)
\end{app_theorem}

Therefore, it is obvious that given any vector field $\mathbf{V} \in L^p(\Omega)^d$ and scalar field $A\in\mathbb{R}_{(0,\, 1]}(\Omega)$ on a bounded domain $\Omega\subset\mathbb{R}^d$ with smooth boundary $\partial \Omega$. If $\mathbf{V}$ satisfies $\nabla \cdot \mathbf{V} = 0$, according to Theorem~\ref{app_thm: repre_div_free}, there exists a $\widetilde{\Psi}$ such that:
\begin{align}
    \mathbf{V} & \leftrightharpoons \nabla \times \widetilde{\Psi} \nonumber \\
    & \leftrightharpoons \nabla \times \left(A \, (\widetilde{\Psi} / A) \right) \nonumber \\
    & \leftrightharpoons \nabla \times (A \, \Psi) \label{app_eq: thm_proof_v} \\ 
    & \leftrightharpoons \nabla A \times \Psi + A \, \nabla \times \Psi \nonumber \\
    & \leftrightharpoons \nabla A \times \Psi + A \, \overline{\mathbf{V}},\, \label{app_eq: proof_relation_v}
\end{align}
where Eq.~(\ref{app_eq: thm_proof_v}) corresponds to Eq.~(\ref{app_eq: repre_v}), and is the explicit expression for the implementation of ``anomaly-encoded'' velocity vector field $\mathbf{V}$ in this paper. $\overline{\mathbf{V}}$ in Eq.~(\ref{app_eq: proof_relation_v}) refers to the ``anomaly-free'' velocity field as defined in Eq.~(\ref{eq: relation}) in the main manuscript where the relation between $\mathbf{V}$ and $\overline{\mathbf{V}}$ it is denoted as $\mathbf{V} = A \diamond \overline{\mathbf{V}}$.

\end{proof}

\section{Theorem \ref{thm: repre_psd}: Anomaly-decomposed Symmetric PSD Tensor Representation}
\label{app: proof_psd}

{\it
\noindent For any $n\times n$ symmetric PSD tensor $\mathbf{D}$ and $A \in \mathbb{R}_{(0,\, 1]}(\Omega)$, there exist an upper triangular matrix with zero diagonal entries, $\mathbf{B} \in \mathbb{R}^{\frac{n(n-1)}{2}}$, and a non-negative diagonal matrix, $\boldsymbol{\Lambda} \in SD(n)$, satisfying:
\begin{equation}
\mathbf{D} = \mathbf{U}\, (A \, \boldsymbol{\Lambda}) \, \mathbf{U}^T,\quad \mathbf{U} = exp(\mathbf{B} - \mathbf{B}^T) \in SO(n).
\label{app_eq: repre_psd}
\end{equation}
Conversely, for $\forall A \in \mathbb{R}_{(0,\, 1]}(\Omega)$, $\forall \mathbf{B} \in \mathbb{R}^{\frac{n(n-1)}{2}}$, and any diagonal matrix with non-negative diagonal entries, $\boldsymbol{\Lambda} \in SD(n)$, Eq.~(\ref{app_eq: repre_psd}) results in a symmetric PSD tensor, $\mathbf{D}$.  
}

\begin{proof}
The above Theorem is a Corollary of the following Theorem \ref{app_thm: repre_d}, which is originally introduced by Liu et al. \cite{liu2021yeti}. 

\begin{app_theorem}[Symmetric PSD Tensor Representation by Spectral Decomposition]
For any tensor $\mathbf{D}  \in PSD(n)$, there exists an upper triangular matrix with zero diagonal entries, $\mathbf{B} \in \mathbb{R}^{\frac{n(n-1)}{2}}$, and a diagonal matrix with non-negative diagonal entries, $\boldsymbol{\Lambda} \in SD(n)$, satisfying: 
\begin{equation}
\mathbf{D} = \mathbf{U}\, \boldsymbol{\Lambda} \, \mathbf{U}^T,\quad \mathbf{U} = exp(\mathbf{B} - \mathbf{B}^T) \in SO(n).
\label{app_eq: repre_d} 
\end{equation}
Conversely, for any upper triangular matrix with zero diagonal entries, $\mathbf{B} \in \mathbb{R}^{\frac{n(n-1)}{2}}$, and any diagonal matrix with non-negative diagonal entries, $\boldsymbol{\Lambda} \in SD(n)$, Eq.~(\ref{app_eq: repre_d}) results in a symmetric PSD tensor, $\mathbf{D}\in PSD(n)$. 
\label{app_thm: repre_d}

\noindent (For detailed proof of Theorem \ref{app_thm: repre_d}, please refer to Supp.~B in \cite{liu2021yeti}.)
\end{app_theorem}

Therefore, $\forall \, n\times n$ symmetric PSD tensors $\mathbf{D}$ and $\forall \, A \in \mathbb{R}_{(0,\, 1]}(\Omega)$, there exists a $\mathbf{B} \in \mathbb{R}^{\frac{n(n-1)}{2}}$, and a $\widetilde{\boldsymbol{\Lambda}} \in SD(n)$, such that:
\begin{align}
\mathbf{D} & \leftrightharpoons \mathbf{U}\, \widetilde{\boldsymbol{\Lambda}} \, \mathbf{U}^T \nonumber \nonumber \\
& \leftrightharpoons \mathbf{U}\,\left( A \, (\widetilde{\boldsymbol{\Lambda}} / A) \right)\, \mathbf{U}^T \nonumber \\
& \leftrightharpoons \mathbf{U}\,(A \, \boldsymbol{\Lambda})\, \mathbf{U}^T \nonumber \label{app_eq: thm_proof_d} \\
& \leftrightharpoons A \, \mathbf{U}\, \boldsymbol{\Lambda} \, \mathbf{U}^T \\
& \leftrightharpoons A \, \overline{\mathbf{D}},\quad \mathbf{U} = exp(\mathbf{B} - \mathbf{B}^T) \in SO(n),\,
\label{app_eq: proof_relation_d}
\end{align}
where Eq.~(\ref{app_eq: thm_proof_d}) corresponds to Eq.~(\ref{app_eq: repre_psd}), and is the explicit expression for the implementation of ``anomaly-encoded'' diffusion tensor field $\mathbf{D}$ in this paper. $\overline{\mathbf{D}}$ in Eq.~(\ref{app_eq: proof_relation_d}) refers to the ``anomaly-free'' diffusion tensor field as defined in Eq.~(\ref{eq: relation}) in the main manuscript where the relation between $\mathbf{D}$ and $\overline{\mathbf{D}}$ it is denoted as $\mathbf{D} = A \circ \overline{\mathbf{D}}$.

\end{proof}

%%%%%%%%%%%%%%%%%%%%%%%%%%%%%%%%%%

\section{\texttt{PyTorch} Deterministic-Stochastic Advection-Diffusion PDE Toolkit: Numerical Derivations}
\label{app: numericals}
Our stochastic advection-diffusion PDE toolkit is designed to solve, separately or jointly, deterministically or stochastically, advection and diffusion PDEs in 1D/2D/3D. 
\begin{align}
\frac{\partial C({\mathbf{x}}, t)}{\partial t} =  \frac{\partial C({\mathbf{x}}, t)}{\partial t}\bigg |_{\text{adv.}} +  \frac{\partial C({\mathbf{x}}, t)}{\partial t}\bigg |_{\text{diff.}} +  \frac{\partial C({\mathbf{x}}, t)}{\partial t}\bigg |_{\text{sto.}} &= \underbrace{-  \nabla \left({\mathbf{V}}({\mathbf{x}}) \cdot C({\mathbf{x}}, t) \right)}_{\text{Fluid flow}} + \underbrace{\nabla \cdot \left({\mathbf{D}}({\mathbf{x}})\, \nabla C({\mathbf{x}}, t)\right)}_{\text{Diffusion}} + \underbrace{\sigma(\mathbf{x}) \partial W(\mathbf{x},\, t)}_{\substack{\text{Model}\\\text{Uncertainty}}} \nonumber \\
(\nabla \cdot \mathbf{V} = 0) & = \underbrace{- {\mathbf{V}}({\mathbf{x}})\cdot\nabla C({\mathbf{x}}, t)}_{\text{Incompressible flow}} + \underbrace{\nabla \cdot \left({\mathbf{D}}({\mathbf{x}})\, \nabla C({\mathbf{x}}, t)\right)}_{\text{Diffusion}} + \underbrace{\sigma(\mathbf{x}) \partial W(\mathbf{x},\, t)}_{\substack{\text{Model}\\\text{Uncertainty}}} \,.
\label{app_eq: adv_diff}  
\end{align}
One can choose to model the velocity field as a constant, a general vector field, or a divergence-free vector field (for incompressible flow). Furthermore, the diffusion field can be modeled as a constant, a non-negative scalar field, or a symmetric positive semi-definite (PSD) tensor field. By controlling$\sigma(\mathbf{x})$, one can specify the variance the Brownian motion $W(\mathbf{x},\, t)$, and thus the levels of uncertainty of the entire system. The toolkit is implemented as a custom \texttt{torch.nn.Module} subclass, such that one can directly use it as a (stochastic) advection-diffusion PDE solver for data simulation or easily wrap it into DNNs or numerical optimization frameworks for inverse (stochastic) PDE problems, i.e., parameters estimation.

%%%%%%%%%%%%%%%%%%%%%%%%%

\subsection{Advection and Diffusion Computation}
\label{sec: impl_adv_diff}
The computation of advection and diffusion and their stability analysis are similar to \texttt{YETI} \cite{liu2021yeti}, please refer to Supp.~C.1-C.2 in \cite{liu2021yeti} for a detailed discussion.

%%%%%%%%%%%%%%%%%%%%%%%%%%%%%%%%%

\subsection{(Stochastic) Numerical Integration}  
As introduced above, after discretizing all the spatial derivatives on the right side of Eq.~(\ref{app_eq: adv_diff}), we obtain a system of ordinary differential equations (ODEs), which can be solved by numerical integration%\footnote{Gwynne  A. Evans, Jonathan M. Blackledge, and Peter  D.Yardley. Numerical Methods for Partial Differential Equations. {\it{Springer London, 2000}}}.
~\cite{evans2000pde}. % To-be-added for arXiv version. 

If the system is modeled stochastically, the stochastic term, $\sigma(\mathbf{x}) \partial W(\mathbf{x},\, t)$, should be included during the integration.

\begin{definition}[Brownian motion] A stochastic process ($W_t$) such that (1) $W_0 = 0$; (2) $(W_t  - W_s) \sim \mathcal{N}(0, t-s), ~\forall t \geq s \geq 0$; (3) For all disjoint time interval pairs $[t_1, t_2]$, $[t_3, t_4]$ $(t_1< t_2 \leq t_3 \leq t_4)$, the increments $W_{t_4} - W_{t_3}$ and $W_{t_2} - W_{t_1}$ are independent random variables.
\end{definition}

We model the advection-diffusion process via a stochastic PDE (SPDE), where $\sigma$ $(\sigma(\mathbf{x}) \in \mathbb{R})$ denotes the variance of the Brownian motion 
$W_t\, (W(\mathbf{x},\, t)\in \mathbb{R})$~\cite{Guerra2009brownian} and represents the epistemic uncertainty for the dynamical system. With this additional stochastic term, the existence and uniqueness of the solution to Eq.~(\ref{app_eq: adv_diff}) still holds:

\begin{theorem}[Existence and uniqueness of SPDE~\cite{oksendal2010stochastic,Guerra2009brownian,Demian2014probreg}]If the coefficients of the stochastic partial differential equation Eq.~(\ref{app_eq: adv_diff}) with initial condition, satisfy the spatially-varying Lipschitz condition
\begin{align}
& |\mathbf{V}(\mathbf{x}_1) - \mathbf{V}(\mathbf{x}_2)|^2 + |\mathbf{D}(\mathbf{x}_1) - \mathbf{D}(\mathbf{x}_2)|^2 \nonumber \\
& \qquad\qquad + |\sigma(\mathbf{x}_1) - \sigma (\mathbf{x}_2)|^2 \leq K|\mathbf{x}_1 - \mathbf{x}_2|^2,
\end{align}
and the spatial growth condition 
\begin{equation}
|\mathbf{V}(\mathbf{x})|^2 +|\mathbf{D}(\mathbf{x})|^2 + |\sigma(\mathbf{x})|^2 \leq K ( 1 + \mathbf{x}^2 ),
\end{equation}
then there is a continuous adapted solution $C({\mathbf{x}}, t)$ satisfying the $L^2$ bound. Moreover, if $C({\mathbf{x}}, t)$ and $\widetilde{C}({\mathbf{x}}, t)$ are both continuous solutions satisfying the $L^2$ bound, then
\begin{equation}
P\left(C({\mathbf{x}}, t) = \widetilde{C}({\mathbf{x}}, t) \text{ for all } t \in [0, \, T]\right) = 1.
\end{equation}
\label{app_thm: spde}
\end{theorem}

For a detailed discussion regarding Theorem~\ref{app_thm: spde}, please refer to~\cite{oksendal2010stochastic,Guerra2009brownian,Demian2014probreg}. During implementation, we use the Euler-Maruyama scheme \cite{Kloeden1992Stochastic,kong2020sde}, and the discretized version of the stochastic term in Eq.~(\ref{app_eq: adv_diff}) can therefore be written as $\sigma(\mathbf{x}) \, W(\mathbf{x},\, t) / \sqrt{\Delta t}$, where $W \sim \mathcal{N}(0,\, 1)$.

We then use the RK45 method to advance in time ($\delta t$) to predict $\widehat{C}^{t+\delta t}$. Note when the input mass transport time-series has relatively large temporal resolution ($\Delta t$), the chosen $\delta t$ should be smaller than $\Delta t$ to satisfy the stability conditions (Supp.~\ref{sec: impl_adv_diff}), thereby ensuring stable numerical integration.

\section{Normal-Abnormal Brain Advection-Diffusion Dataset}
\label{app: ixi}

%%%%%%%%%%%%%%%%%%%%%%%%%%% 
%\input{sub/appendix/fig/v_generator}
%\input{sub/appendix/fig/d_generator}
%\input{sub/appendix/fig/anom_generator}
%\input{sub/appendix/fig/movie_generator}
%%%%%%%%%%%%%%%%%%%%%%%%%%%

Our brain advection-diffusion simulation dataset is based on the public \texttt{IXI} brain dataset\footnote{Available for download: \url{http://brain-development.org/ixi-dataset/}.}, from which we use 200 patients with complete collections of T1-/T2-weighted images, magnetic resonance angiography (MRA) images, and diffusion-weighted images (DWI) with 15 directions for the simulation of 3D divergence-free velocity vector and symmetric PSD diffusion tensor fields. All images above are resampled to isotropic spacing ($1\,mm$), rigidly registered intra-subject (according to the MRA image), and brain-extracted using \texttt{ITK}\footnote{Code in \url{https://github.com/InsightSoftwareConsortium/ITK}.}. 

In general, the generation of the ``anomaly-free'' velocity vector fields and ``anomaly-free'' diffusion tensor fields are of the same procedures as in~\cite{liu2021yeti} (Please refer to Supp.~D.1-D.2 in \cite{liu2021yeti}). Here, Supp.~\ref{app_sec: anomaly} provides the anomaly value fields ($A$) simulation proposed in this paper, and Supp.~\ref{app_sec: movie} introduces the time-series simulation for normal-abnormal advection-diffusion processes.

%%%%%%%%%%%%%%%%%%%%%%%%%%% 
 
\subsection{``Anomaly-encoded'' Velocity Vector and Diffusion Tensor Fields (Fig.~\ref{app_fig:anom_gen})} 
\label{app_sec: anomaly}

\input{sub/appendix/fig/anom_generator}

For each case, we simulate both normal samples and anomaly-encoded samples. For samples treated as normal, we directly use the simulated $\mathbf{V}$ and $\mathbf{D}$ for the concentration time-series simulation. For anomaly encoding, the originally simulated $\mathbf{V}$ and $\mathbf{D}$ are treated as the ``anomaly-free'' fields for generating anomaly-encoded fields, $\overline{\mathbf{V}}$ and $\overline{\mathbf{D}}$, via Eq.~(\ref{eq: relation}). For the generation of the anomaly value field $A$, its spatially varying value (within $[0, 1]$) is determined by a Gaussian with center uniformly sampled across the entire spatial domain of the sample case.

%%%%%%%%%%%%%%%%%%%%%%%%%%% 
 
\subsection{Brain Advection-diffusion Time-series Simulation} % (Fig.~\ref{app_fig: movie_gen})} 
\label{app_sec: movie}

For each brain advection-diffusion sample, the initial concentration state is assumed to be given by the MRA image with intensity ranges rescaled to $[0,\,1]$. %\mnl{What motivates this scale? Is it based on what you expect to see?} 
Time-series (length $N_T = 40$, interval $\Delta t = 0.1\,s$) are then simulated given the computed divergence-free velocity fields and symmetric PSD diffusion tensor fields by our advection-diffusion PDE solver (Supp.~\ref{app: numericals}). Thus the simulated dataset includes 800 brain advection-diffusion time-series (4 time-series for each of the 200 subjects, based on the four combinations of the simulated two velocity fields and two diffusion fields).%\mnl{How do you get 4 time-series form the four different velocity/diffusion combinations?} 
 
%%%%%%%%%%%%%%%%%%%%%%%%%%% 

%\input{sub/appendix/app_results} 
%\input{sub/appendix/math}

%\input{sub/appendix/main_empty}

%\input{sub/appendix/main_empty} % avoid refs. in supp. appear in main body refs. for main text submission.

\end{document}